\documentclass[10pt,twocolumn,letterpaper]{article}

\usepackage{iccv}
\usepackage{times}
\usepackage{epsfig}
\usepackage{graphicx}
\usepackage{amsmath}
\usepackage{amssymb}

\usepackage[font=small,skip=4pt]{caption}
\usepackage{subcaption}
\usepackage{booktabs}
\usepackage{xspace}
\usepackage[english]{babel}
\usepackage{array}

\newcolumntype{P}[1]{>{\centering\arraybackslash}p{#1}}
\newcolumntype{C}[1]{>{\centering\arraybackslash}m{#1}}

\makeatletter
\DeclareRobustCommand\onedot{\futurelet\@let@token\@onedot}
\def\@onedot{\ifx\@let@token.\else.\null\fi\xspace}
\def\eg{\emph{e.g}\onedot} 
\def\ie{\emph{i.e}\onedot}

\def\etal{\emph{et al}\onedot}
\DeclareMathAlphabet{\mathcal}{OMS}{cmsy}{m}{n}
\newcommand{\ve}[1]{\ensuremath{\mathbf{#1}}} %
\newcommand{\set}[1]{\ensuremath{\mathcal{#1}}} %
\usepackage[dvipsnames]{xcolor}

\usepackage[pagebackref=true,breaklinks=true,letterpaper=true,colorlinks,bookmarks=false]{hyperref}

\iccvfinalcopy 

\def\iccvPaperID{7580} 

\ificcvfinal\fi

\begin{document}

\title{A Latent Transformer for Disentangled Face Editing in Images and Videos}

\author{Xu Yao$^{1,2}$, Alasdair Newson$^{1}$, Yann Gousseau$^{1}$, Pierre Hellier$^{2}$ \\
$^{1}$ LTCI, T\'el\'ecom Paris, Institut Polytechnique de Paris, France\\
$^{2}$ InterDigital R\&I, 975 avenue des Champs Blancs, Cesson-S\'evign\'e, France\\
{\tt\small \{xu.yao,anewson,yann.gousseau\}@telecom-paris.fr, Pierre.Hellier@InterDigital.com}
}

\twocolumn[{%
\renewcommand\twocolumn[1][]{#1}%
\maketitle
\small
\setlength{\tabcolsep}{0pt}
\centering
\begin{tabular}{P{0.165\linewidth}P{0.165\linewidth}P{0.165\linewidth}P{0.165\linewidth}P{0.165\linewidth}P{0.165\linewidth}}
Original&Projected& + Smile & + Bangs & + Arched Eyebrows & + Age
\\
\multicolumn{6}{c}{\includegraphics[width=0.99\linewidth]{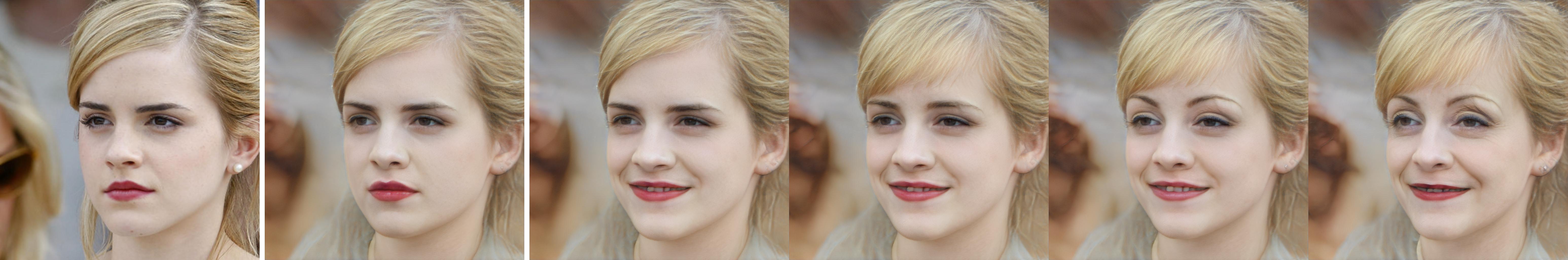}}
\\
&& - Age & + Smile & + Beard & + Eyeglasses
\\
\multicolumn{6}{c}{\includegraphics[width=0.99\linewidth]{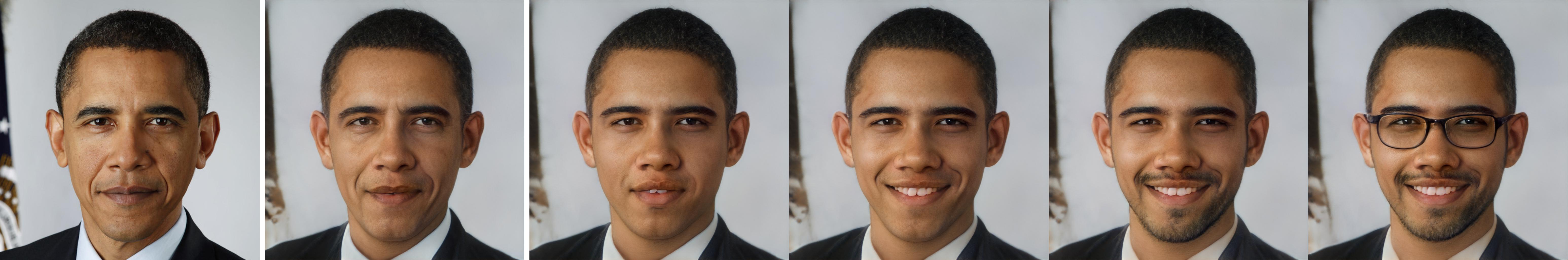}}
\end{tabular}
\label{teaser}
\captionof{figure}{We project real images to the latent space of a StyleGAN generator and achieve sequential disentangled attribute editing on the encoded latent codes. From the original and the projected image, we can edit sequentially a list of attributes such as: `smile', `bangs', `arched eyebrows', `age', `beard' and `eyeglasses'. All results are obtained at resolution $1024 \times 1024$. %
\\}%
}]

\ificcvfinal\thispagestyle{empty}\fi

\begin{abstract}
High quality facial image editing is a challenging problem in the movie post-production industry, requiring a high degree of control and identity preservation. Previous works that attempt to tackle this problem may suffer from the entanglement of facial attributes and the loss of the person's identity. Furthermore, many algorithms are limited to a certain task. To tackle these limitations, we propose to edit facial attributes via the latent space of a StyleGAN generator, by training a dedicated latent transformation network and incorporating explicit disentanglement and identity preservation terms in the loss function. We further introduce a pipeline to generalize our face editing to videos. Our model achieves a disentangled, controllable, and identity-preserving facial attribute editing, even in the challenging case of real (i.e., non-synthetic) images and videos. We conduct extensive experiments on image and video datasets and show that our model outperforms other state-of-the-art methods in visual quality and quantitative evaluation. Source codes are available at \url{https://github.com/InterDigitalInc/latent-transformer}.
\end{abstract}

\section{Introduction}

Facial attribute editing is a crucial task for photo retouching or in the film post-production industry. For example, many actors are retouched for beautification or other special makeup effects. For such tasks, it is highly desirable for artists to be able to control a facial attribute without affecting other informations. Consequently, a face editing method should rely on disentangled attributes and permit identity-preserving manipulations.  

Earlier works based on deep learning focus on encoder-decoder based architectures \cite{choi2018stargan,huang2018multimodal}. Despite the improvements in quality of recent results, these approaches are limited in resolution and generate noticeable artifacts on high resolution images. Therefore, they are not appropriate for high quality video editing. In addition, these methods are difficult to control, because the modification of one facial attribute tends to modify other attributes.

Recently, generative networks have shown impressive progress in high quality image synthesis \cite{biggan,karras2018progressive,karras2019style,karras2019analyzing}. Studies show that moving latent codes along certain directions in the latent space of generative models can result in variations of visual attributes in the corresponding generated images \cite{abdal2020image2stylegan++,collins2020editing,shen2019interpreting,abdal2020styleflow,wu2020stylespace}. These assume that for a binary attribute, there exists a hyper-plane in the latent space which divides the data into two groups. However, this hypothesis has several limitations. Firstly, successful manipulations can only be achieved in well disentangled and linearized latent spaces. Although the latent space is disentangled compared to the image space, we show in this paper that achieving facial attribute manipulation with linear transformations is a very strong and limiting hypothesis. %
Furthermore, since these methods are trained on synthetic images (generated from random points in the latent space), their performance on real images (natural, ``in-the-wild'' photos) is less satisfying. This is an often ignored, but critical, problem.

In this work, we tackle the problem of editing facial attributes on real images. To address the aforementioned limitations, we propose a transformation network to navigate the latent space of the generative model. We project real images to the latent space of the state-of-the-art image generator StyleGAN and train our model on the projected latent codes. The transformation network generates disentangled, identity-preserving and controllable attribute editing results on real images.
These key advantages allow us to extend our method to the case of videos, where stability and quality are of crucial importance. For this, we introduce a pipeline which achieves stable and realistic facial attribute editing on high resolution videos.

\textbf{Our contributions} can be summarized as follows. We propose a latent transformation network for facial attribute editing, achieving disentangled and controllable manipulations on real images with good identity preservation. Our method can carry out efficient sequential attribute editing on real images. We introduce a pipeline to generalize the face editing to videos and generate realistic and stable manipulations on high resolution videos.

The rest of the paper is organized as follows: In Section 2 we summarize the related works in facial attribute editing, disentangled representation and video editing. Section 3 presents our latent transformation network and the training details.
In Section 4 we present the experimental results of disentangled attribute editing on real images, and provide qualitative and quantitative comparisons with state-of-the-art methods. We further present results of sequential attribute editing on real images and give an ablation study on the choice of loss compositions.
In Section 5 we introduce the pipeline to apply facial attribute editing on videos and show experimental results on video sequences. We conclude the paper in Section 6.

\section{Related works}
\textit{Facial Attribute Editing.} Previous works regarding facial attributes are extensive and mainly focus on images of limited resolution. An optimization-based approach by Upchurch \etal \cite{upchurch2017deep} showed that it is possible to achieve semantic transformations such as aging or adding facial hair by interpolating deep features in a pre-trained feature space. Another type of approach train feed-forward models for the attribute editing task. Attribute2image \cite{yan2016attribute2image} proposed to train a Conditional Variational Auto-Encoder to generate attribute-conditioned images. With the success of generative networks in image synthesis, a number of studies \cite{choi2018stargan,he2019attgan,lample2017fader,liu2019stgan,wu2020leed} explored the possibility of training auto-encoders using adversarial learning. FaderNet \cite{lample2017fader} and StarGAN \cite{choi2018stargan} proposed to disentangle different attributes in the latent space of auto-encoder and generate the output image conditioned on the target attributes. AttGAN \cite{he2019attgan} and STGAN \cite{liu2019stgan} enhanced the flexible translation of attributes to improve the image quality by relaxing the strict constraints on the target attributes. Several studies investigate different possibilities to tackle high resolution images. CooGAN \cite{chen2020coogan} proposed a patch-based local-global framework to process HR images in patches. Observing the great progress of generative networks in high quality image synthesis, Viazovetskyi \etal \cite{viazovetskyi2020stylegan2} trained the pix2pixHD model \cite{wang2018pix2pixHD} for single attribute editing with the synthetic images generated by StyleGAN2 \cite{karras2019analyzing}.

\textit{Disentangled Representations.} In the paper of StyleGAN \cite{karras2019style}, the authors examined the effects of mixing two latent codes on the generated image (which is referred as style mixing), and found that each subset controls meaningful high-level attributes of the image. Inspired by this, some studies have attempted to explore disentangled representations in the latent space of generative networks, especially StyleGAN. One optimization-based method, Image2StyleGAN++ \cite{abdal2020image2stylegan++}, carried out local editing along with 
global semantic edits on images by applying masked interpolation on the activation features of StyleGAN. Collins \etal \cite{collins2020editing} performed a k-means clustering on the activations of StyleGAN and detected a disentanglement of semantic objects, which enables further local semantic editing on the generated image. For high level semantic edits, Ganalyze \cite{goetschalckx2019ganalyze} learned a manifold in the latent space of BigGAN \cite{biggan} to generate images of different memorability. InterFaceGAN \cite{shen2019interpreting} proposed to learn a hyper-plane for a binary classification in the latent space, which one can use to manipulate the target facial attribute by simple interpolation. Following their work, StyleSpace \cite{wu2020stylespace} carried out a quantitative study on the latent spaces of StyleGAN \cite{karras2019analyzing} and realized a highly localized and disentangled control of the visual attributes. StyleFlow \cite{abdal2020styleflow} achieved conditional exploration of the latent space by training conditional normalizing flows. StyleRig \cite{tewari2020stylerig} introduced a method to provide a face rig-like control over a pretrained and fixed StyleGAN via a 3D morphable face model. Yao \etal \cite{yao2021learning} proposed to navigate the latent space of StyleGAN in a non-linear manner to achieve disentangled manipulations of facial attributes. To find out the disentangled directions, some researches attempted to analyze the latent space of generative networks using Principal Component Analysis (PCA). PCAAE \cite{pham2020pcaae} presented an PCA auto-encoder, whose latent space is progressively increased during training and results in a separation of intrinsic data attributes into different components. GANSpace \cite{harkonen2020ganspace} performed PCA in the latent space of generative networks, explored the principal directions and discovered interpretable controls. The above-mentioned methods generally focus on manipulations of synthetic images, as it remains a challenge to project real images to the latent space of StyleGAN. Image2StyleGAN used optimization method to project real images to an extended latent space of StyleGAN, but the characteristics of which differ from the original latent space thus not suitable for manipulation. Some recent works \cite{park2020swapping,pidhorskyi2020adversarial,richardson2020encoding,zhu2020domain} trying to train an encoder together with the StyleGAN model. Although the images cannot be perfectly reconstructed, we see the possibility of carrying out attribute editing on real images using the disentanglement characteristics of the StyleGAN latent space.

\textit{Video-based Face Editing.} Recent works on face video synthesis mainly address two problems: 1) generating a face video sequence from a sketch video or a reference image (often referred as face reenactment), and 2) facial attribute editing on videos. Garrido \etal \cite{garrido2014automatic} proposed an image-based reenactment system to achieve face replacement in video. Face2Face \cite{thies2016face2face} presented an approach for real-time facial reenactment of a target video using non-rigid model-based bundling. Averbuch-Elor \etal \cite{averbuch2017bringing} presented a technique to animate a still portrait with a driving video, but limited to mild movements. Kim \etal \cite{kim2018deep} proposed to use generative neural networks for re-animation of portrait videos, which transforms not only the facial expressions but also the full upper body and background. Most of these methods handle only low-quality video shots. A popular open-source deepfake system DeepFaceLab \cite{petrov2020deepfacelab} has attracted much attention. Incorporating productivity tools such as manual face detection and landmark extraction tool, their pipeline generates high fidelity face swapping results on videos. To realize facial attribute editing directly on videos, Rav-Acha \etal \cite{rav2008unwrap} suggested to convert video frames to an “unwrap mosaic”, paint and re-render the mosaic to videos. Despite satisfying results, computing the mosaic for each video shot is a long process and requires a smooth variation to construct successfully. Duong \etal \cite{duong2019automatic} proposed an approach to generate age-progressed facial images in video sequences using deep reinforcement learning. Many recent works use deep learning techniques to realize facial attribute editing on still images. However, up to now, only a few works have addressed video-based attribute editing problem \cite{zheng2020survey}.

\section{Method}

In this section, we propose a framework to edit faces in real images and videos via the latent space of StyleGAN.

\subsection{Overview}
For a given real image \ve{I}, we assume that we can compute a latent representation \ve{w} $\in$ \set{W} associated to a generator \ve{G}. An inversion method is trained so that $\ve{I} \approx \ve{G(w)}$.
We aim to train a latent transformer \ve{T} in the latent space to edit a single attribute of the projected image \ve{G(w)}. The image synthesized from \ve{T(w)} is denoted by \ve{G(T(w))}. It shares all the attributes with \ve{G(w)} except the target attribute being manipulated. 

To train the latent transformer, we propose a training framework that computes all the losses solely in the latent space \set{W}. Let $\{\ve{a}_1, \ve{a}_2, ..., \ve{a}_N\}$ be a set of image attributes, where $N$ is the total number of considered attributes. For each attribute $\ve{a}_k$, a different $\ve{T}_k$ is trained. 
To predict the attributes from the latent codes we use a latent classifier $\ve{C}: \set{W} \rightarrow \{0,1\}^N$. $\ve{C}$ is pre-trained and then its weights are frozen during the training of $\ve{T}_k$.
We train $\ve{T}_k$ with the following three objectives:

\begin{itemize}
    \item To ensure that $\ve{T}_k$ manipulates attribute $\ve{a}_k$ effectively, we minimize the binary classification loss:
\begin{equation}
\label{eq:class}
\mathcal{L}_{\rm cls} = - y_k\ {\rm log\ }(\ve{p}_k) - (1 - y_k)\ {\rm log\ }(1 - \ve{p}_k),   
\end{equation}
where $\ve{p}_k = \ve{C}(\ve{T}_k(\ve{w}))[k]$ is the probability of the target attribute and $y_k \in \{0, 1\}$ is the desired label.

\item To ensure that other attributes $\ve{a}_i, i\neq k$ remain the same, we apply an attribute regularization term:
\begin{equation}
\label{eq:attr}
\mathcal{L}_{\rm attr} = \sum_{i\neq k}(1-\gamma_{ik})\ \mathbb{E}_{\ve{w}, i}[||\ve{p}_i - \ve{C}(\ve{w})[i]||_2],
\end{equation}
where $\gamma_{ik}$ is the absolute correlation value between $\ve{a}_i$ and the target attribute $\ve{a}_k$, measured on the training dataset. This regularization term is weighted based on correlation to prevent the attributes which are naturally correlated with the target from being over-constrained, \ie `chubby' and `double chin'. 

\item To ensure that the identity of the person is preserved, we further apply a latent code regularization:
\begin{equation}
\label{eq:recon}
\mathcal{L}_{\rm rec} = \mathbb{E}_{\ve{w}}[||\ve{T(w)} - \ve{w}||_2].
\end{equation}

\end{itemize}
The full objective loss can be described as:
\begin{equation}
\label{eq:total}
\mathcal{L} =  \mathcal{L}_{\rm cls} + \lambda_{\rm attr}\mathcal{L}_{\rm attr} + \lambda_{\rm rec}\mathcal{L}_{\rm rec},
\end{equation}
where $\lambda_{\rm attr}$ and $\lambda_{\rm rec}$ are weights balancing each loss.

\begin{figure*}[t]
\centering
\includegraphics[width=0.99\linewidth]{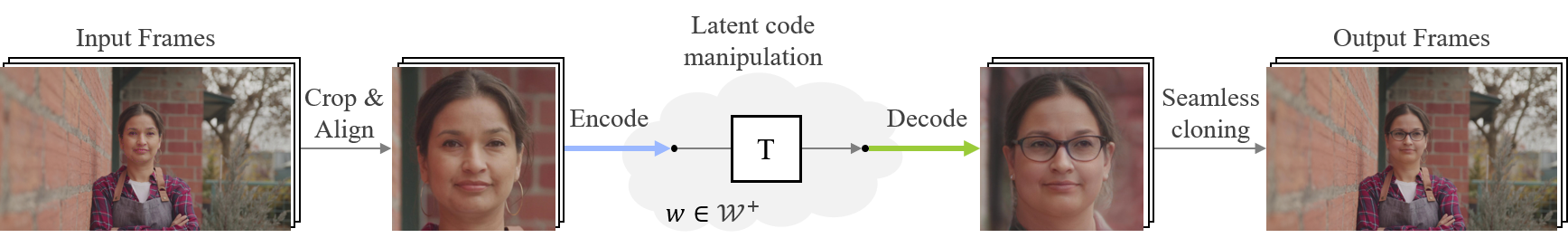}
\caption{\textbf{Video manipulation pipeline.} Each input frame is cropped and aligned to a face image individually. A pretrained encoder \cite{richardson2020encoding} is used to encode the face images to the latent space \set{W^+} of StyleGAN \cite{karras2019analyzing}. The obtained latent codes are processed by the proposed latent transformer \ve{T} to realize the attribute editing. The manipulated latent codes are further decoded by StyleGAN to generate the manipulated face images, which are blended with the original input frames to get the output frames.   
}
\label{video_pipeline}
\end{figure*}
\subsection{Training and models}

To realize attribute editing on real images, we first need to compute the corresponding latent codes in the latent space of StyleGAN. 
Unlike a traditional generative network which feeds random vectors as an input to the first layers, StyleGAN has a different design. The generator takes a constant tensor as input, while each convolution layer output is controlled by style codes via adaptive instance normalization layers \cite{huang2017arbitrary}. A Gaussian random latent code $\ve{z} \in \set{Z}$ is first passed through a mapping network to get an intermediate latent code $\ve{w} \in \set{W}$, which is further specialized by learned affine transforms to style codes \ve{y}. Given a target image \ve{x}, finding the corresponding latent code in \set{W} remains difficult, and the quality of the reconstruction is not fully satisfactory. Image2StyleGAN \cite{abdal2019image2stylegan} go a step further by computing the latent code in an extended latent space \set{W^+}, where latent code \ve{w} controlling each layer may be different, whereas the original setting requires them to be the same. The target image is thus better reconstructed from the latent code obtained in \set{W^+}. 

In our approach, we train the latent transformer \ve{T} in the latent space \set{W^+}, which is specifically designed for the projection of latent codes of real images.
To prepare the training data, we compute the latent codes in \set{W^+} for each image in CelebA-HQ dataset \cite{karras2018progressive}, using a pre-trained StyleGAN encoder proposed by Richardson \etal \cite{richardson2020encoding}. Combined with the annotations for each image, we obtain the ``latent code - label'' pairs as our training data.

\textbf{Latent Classifier.} To predict attributes on the manipulated latent codes, we train an attribute classifier $\ve{C}$ on the ``latent code - label'' pairs. The classifier consists of three fully connected layers with ReLU activations in between. $\ve{C}$ is \textit{fixed} during the training of the latent transformer.

\textbf{Latent Transformer.} Given a latent code $\ve{w} \in \set{W^+}$, the latent transformer \ve{T} generates the direction for a single attribute modification, where the amount of changes is controlled by a scaling factor $\alpha$. The network is expressed with a single layer of linear transformation $f$: 
\begin{equation}
\label{eq:tnet}
    \ve{T(w, \alpha)} = \ve{w} + \alpha \cdot f(\ve{w}).
\end{equation}
During training the scaling factor $\alpha$ is set according to the probability $p$ of the target attribute of the input latent code ($1-p$ for $p < 0.5$, $-p$ for $p > 0.5$). At test time, $\alpha$ can be sampled from $[-1, 1]$ or set beyond this range based on the desired amount of changes.

\subsection{Video manipulation}

In this section, we propose a pipeline which applies the image editing method to the case of videos. The encoding process ensures that the encoded latent codes of two consecutive frames are similar to each other. Therefore, we can reconstruct a face video using the frames projected to the latent space of StyleGAN, which provides the basics for the next manipulation step. Thanks to the stability of our proposed latent transformer, the manipulation does not affect the consistency between the latent codes and generates stable edits on the projected frames. An overview of our proposed pipeline is presented in Figure \ref{video_pipeline}. The pipeline consists of three steps: pre-processing, image editing and seamless cloning.

\noindent \textbf{Pre-processing.} In order to edit the video in the latent space of StyleGAN, we first extract face images from the frames, according to the StyleGAN setting. We crop and align each frame around the face, following the pre-processing step of FFHQ dataset \cite{karras2019style}, on which the StyleGAN is pretrained. For face alignment we detect landmarks independently on each frame using a state-of-the-art method \cite{bulat2017far}. To avoid jitter, we further process the landmarks using optical flow between two consecutive frames and a Gaussian filtering along the whole sequence. All frames are cropped and aligned to have eyes at the center and resized to $1024 \times 1024$.

\noindent \textbf{Image editing.} In this step, we apply our manipulation method on the processed face images. Each frame is encoded to the latent space of StyleGAN using the pre-trained encoder \cite{richardson2020encoding}. The encoded latent codes are processed by the proposed latent transformer to realize the attribute editing. The manipulated latent codes are further decoded by StyleGAN to generate the manipulated face images.

\begin{figure*}[!ht]
\centering
\small
\setlength{\tabcolsep}{0.4pt}
\renewcommand{\arraystretch}{0.5}
\begin{tabular}{P{0.124\linewidth}P{0.124\linewidth}P{0.124\linewidth}P{0.124\linewidth}P{0.124\linewidth}P{0.124\linewidth}P{0.124\linewidth}P{0.124\linewidth}}
    &&&+ Gender&&&+ Beard&
    \\
    \multicolumn{2}{c}{\includegraphics[width=0.248\linewidth]{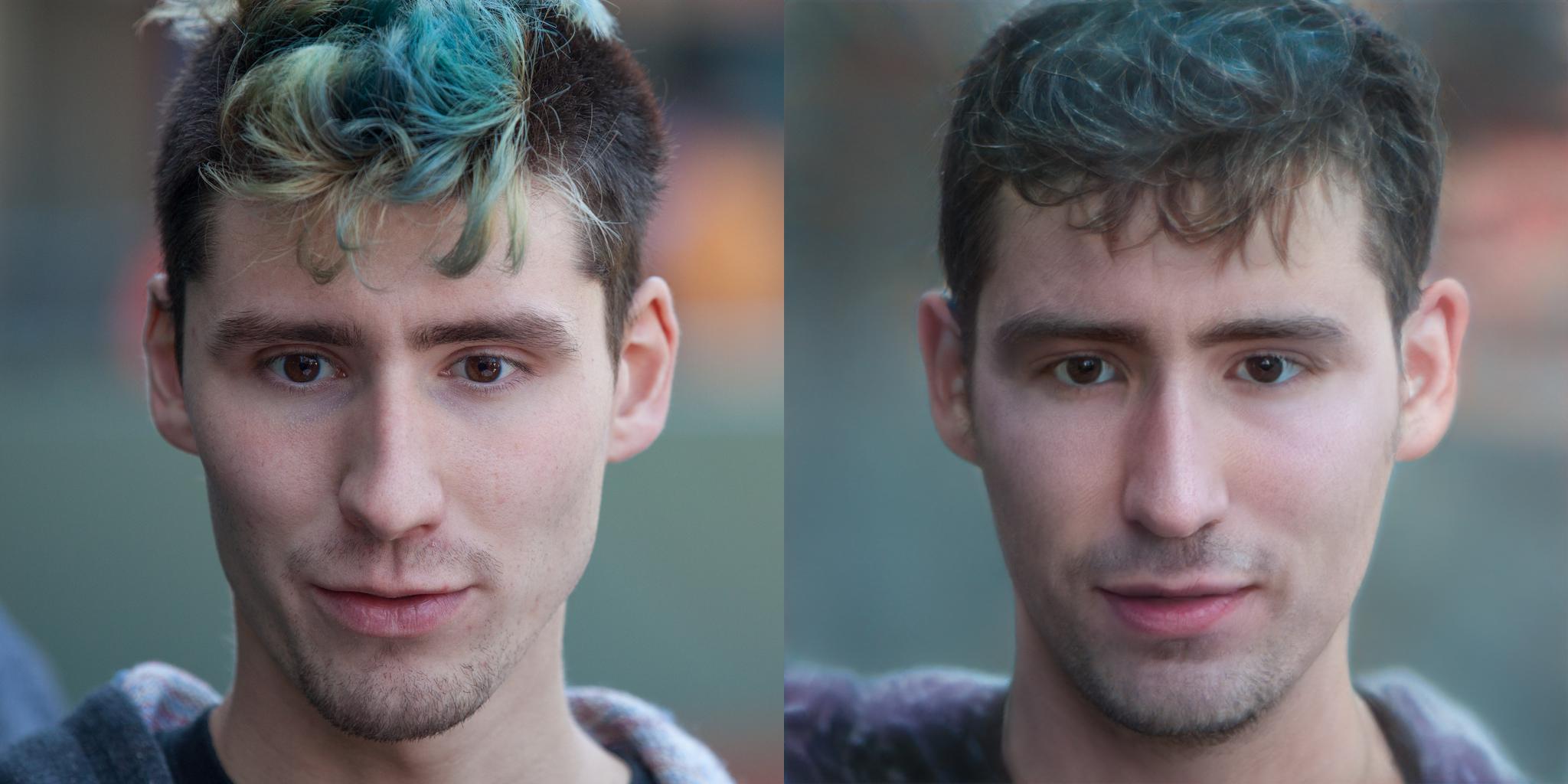}}
    &\multicolumn{3}{c}{\includegraphics[width=0.372\linewidth]{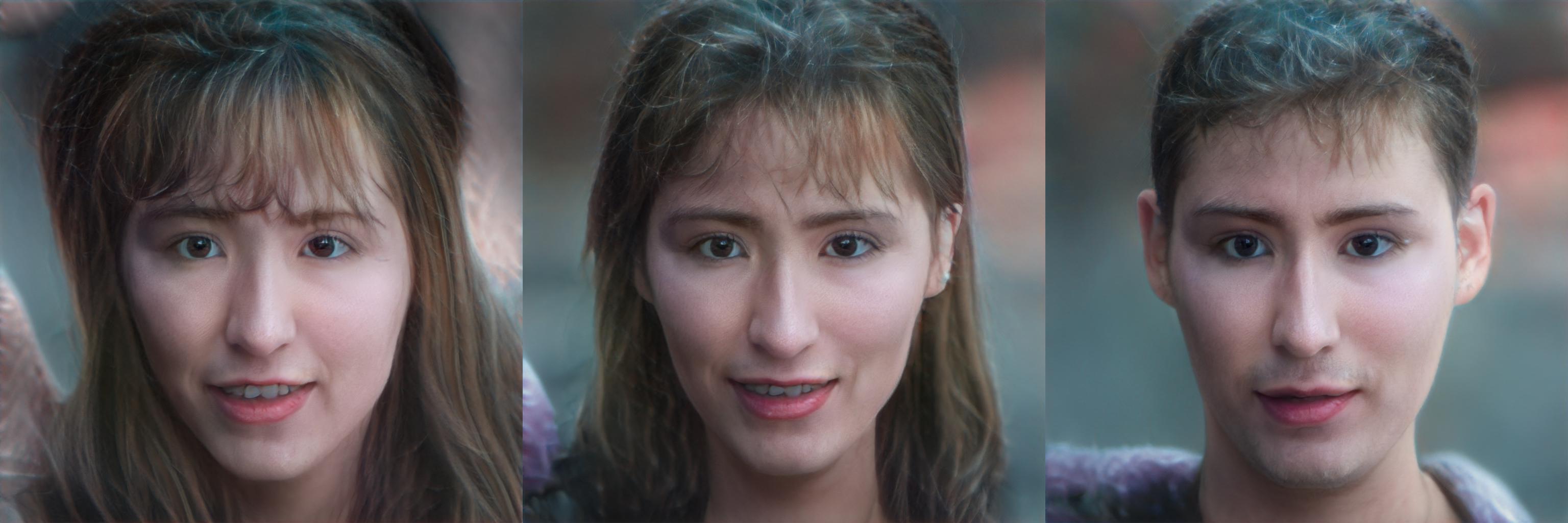}}
    &\multicolumn{3}{c}{\includegraphics[width=0.372\linewidth]{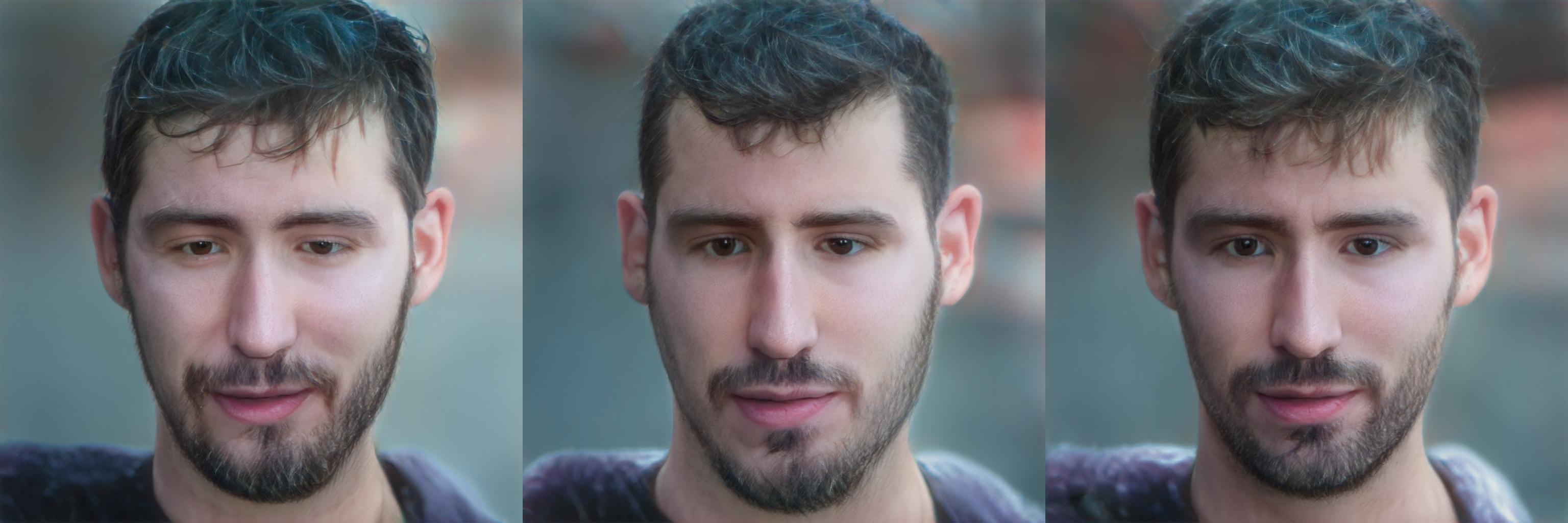}}
    \\
    \multicolumn{2}{c}{\includegraphics[width=0.248\linewidth]{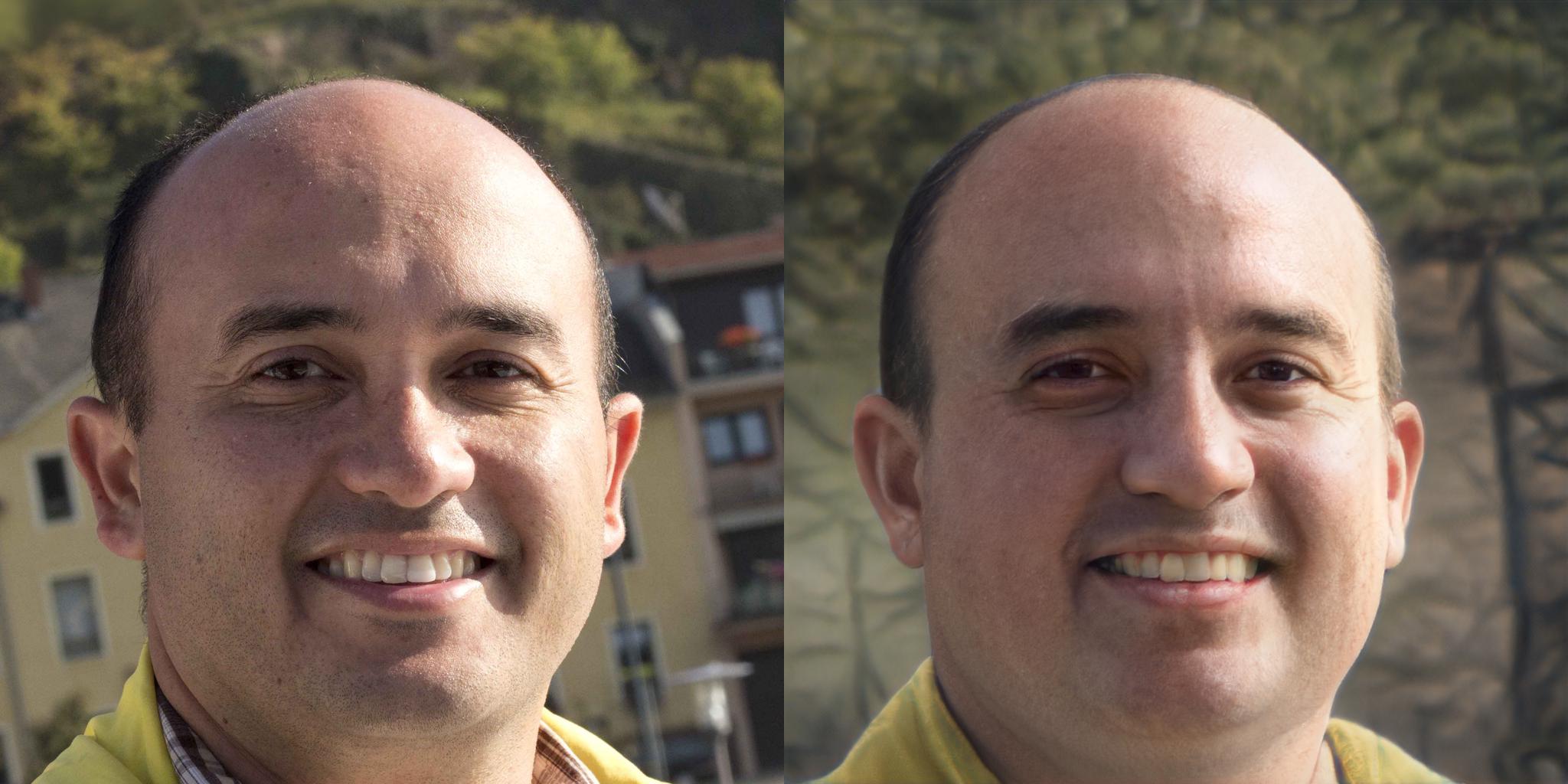}}
    &\multicolumn{3}{c}{\includegraphics[width=0.372\linewidth]{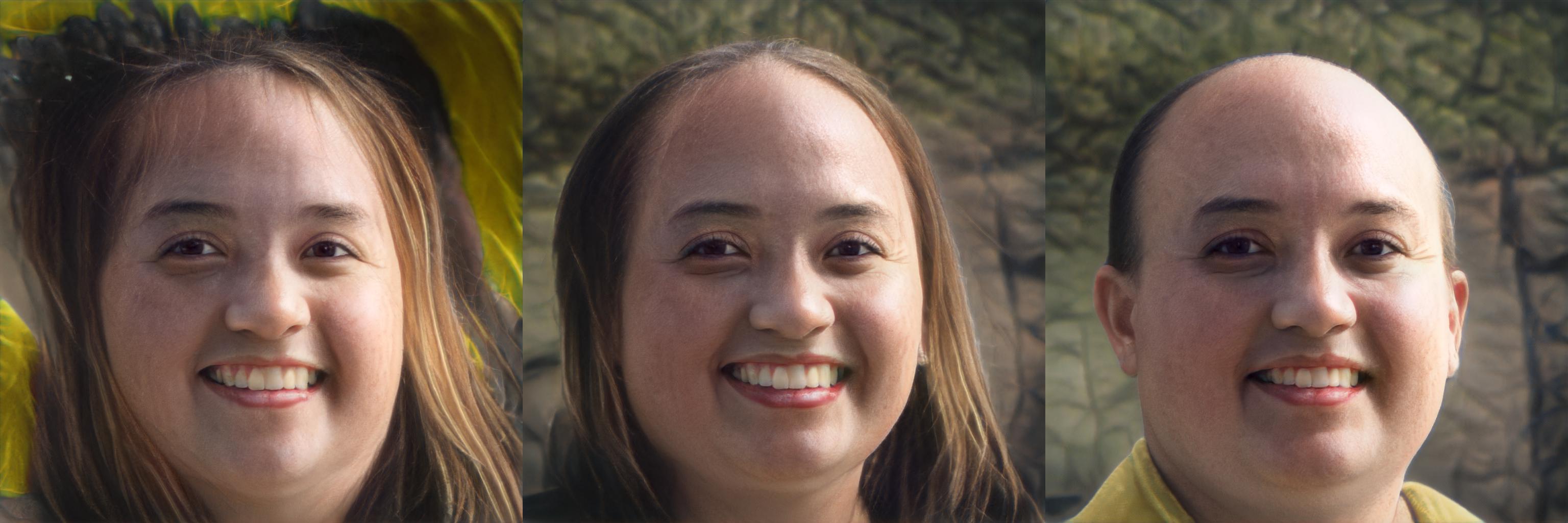}}
    &\multicolumn{3}{c}{\includegraphics[width=0.372\linewidth]{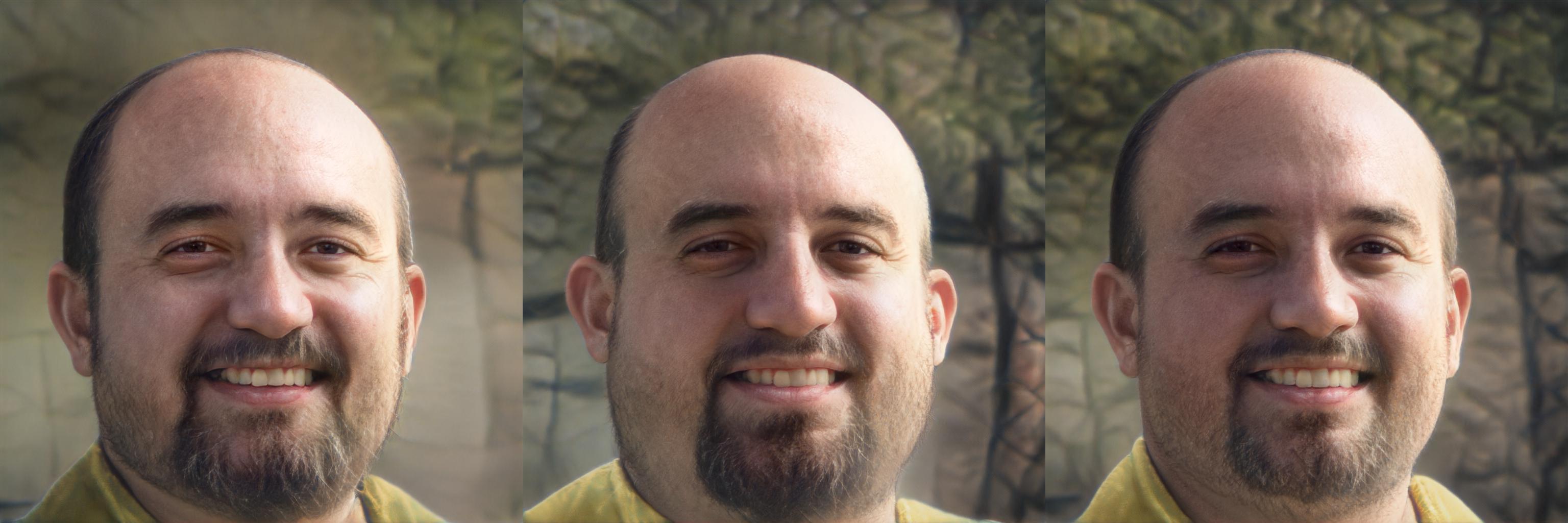}}
    \\
    \hline
    \\
    &&&+/- Age&&&+ Makeup&
    \\
    \multicolumn{2}{c}{\includegraphics[width=0.248\linewidth]{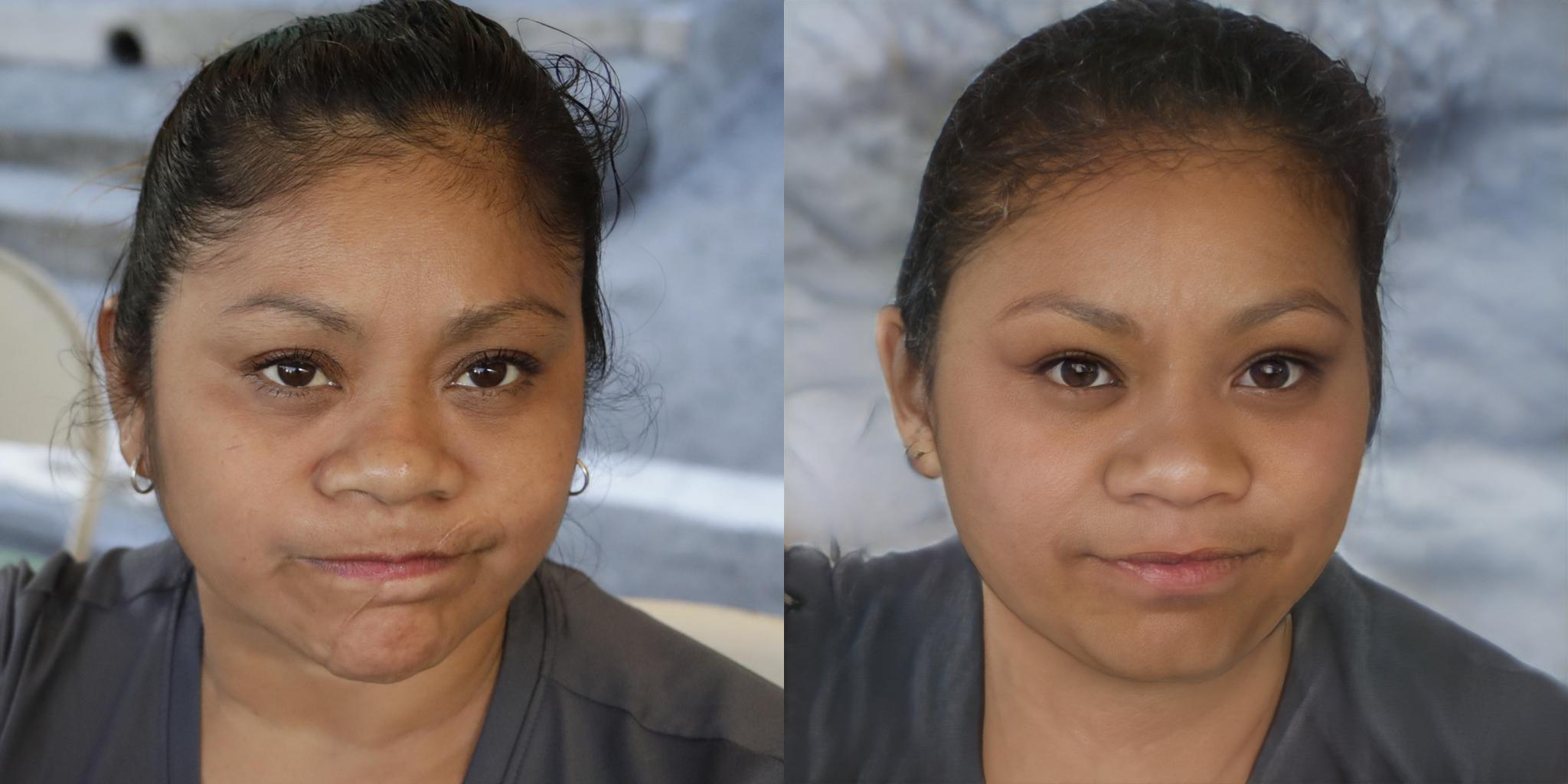}}
    &\multicolumn{3}{c}{\includegraphics[width=0.372\linewidth]{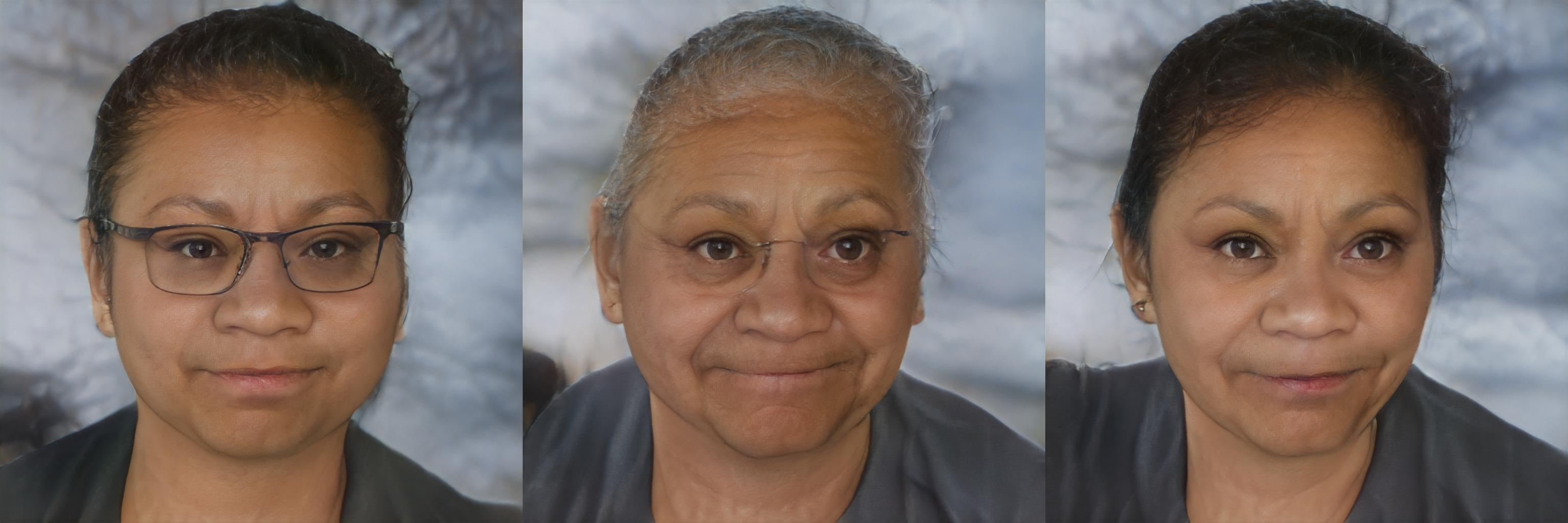}}
    &\multicolumn{3}{c}{\includegraphics[width=0.372\linewidth]{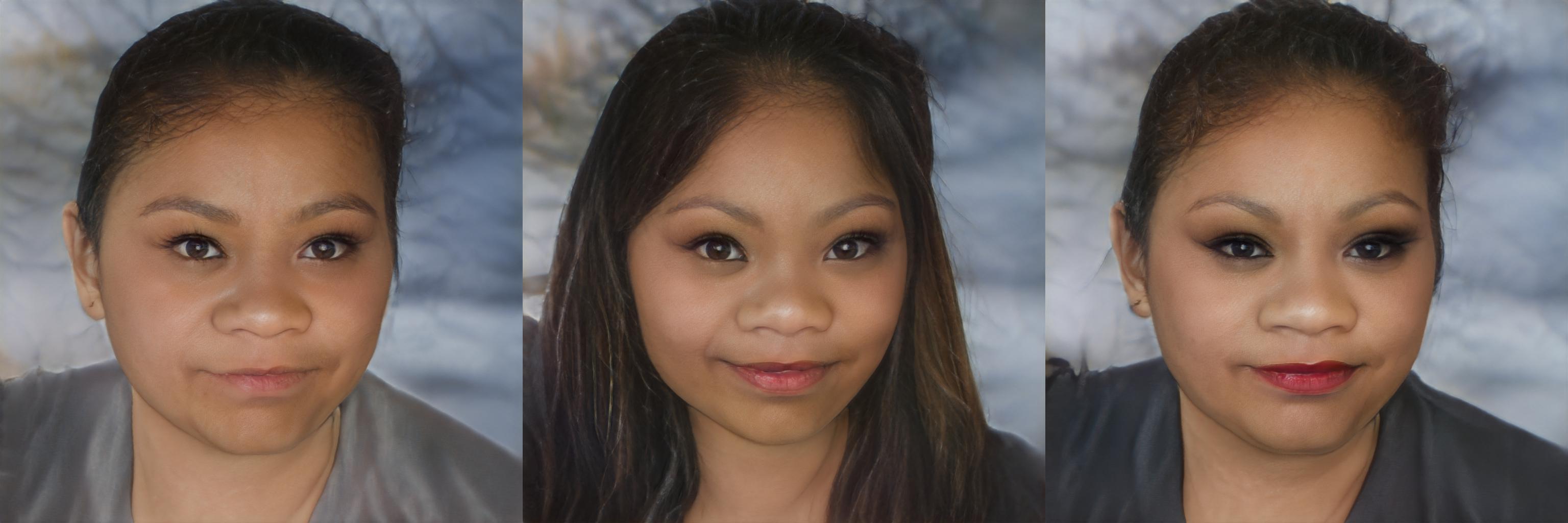}}
    \\
    \multicolumn{2}{c}{\includegraphics[width=0.248\linewidth]{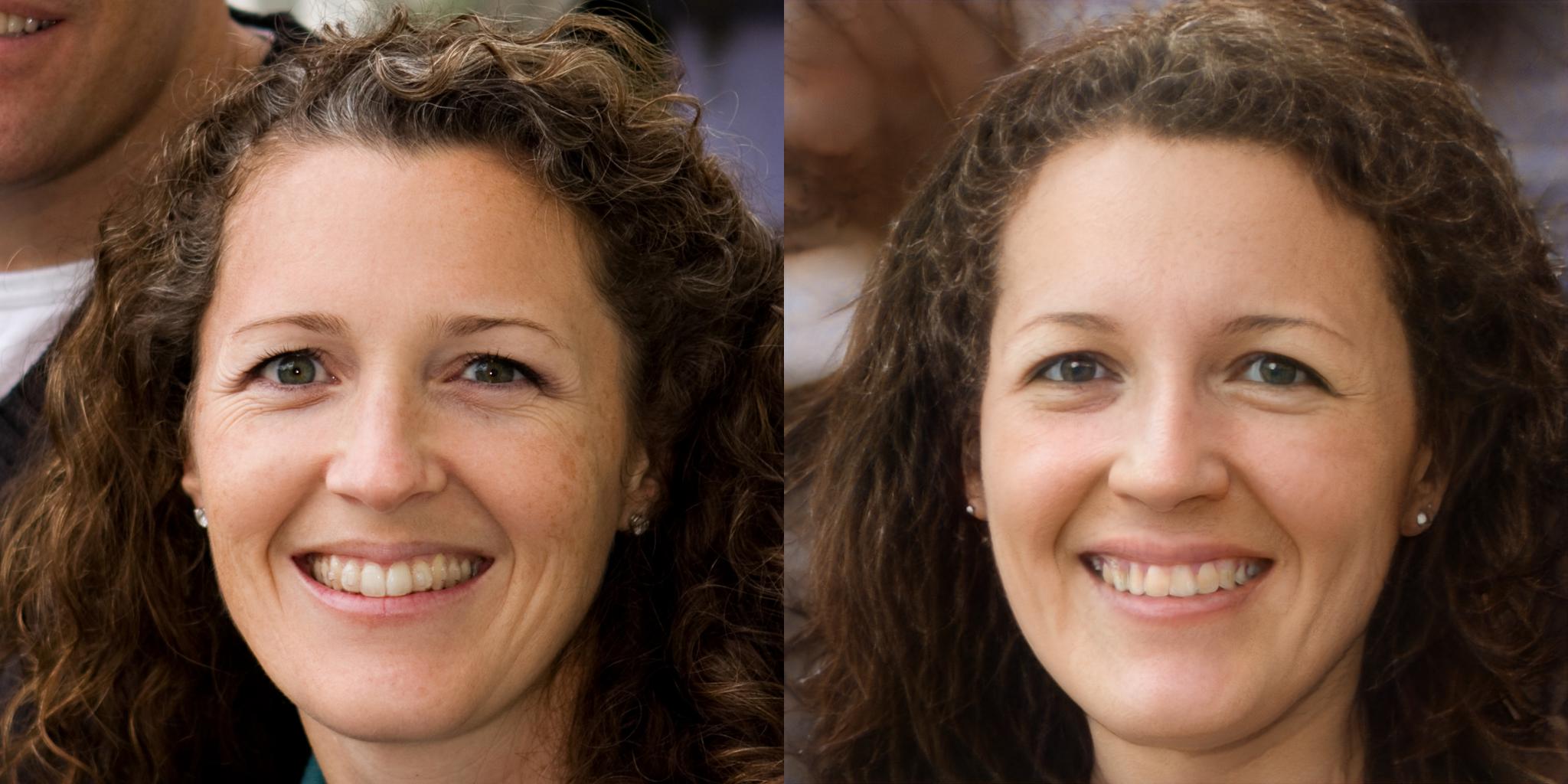}}
    &\multicolumn{3}{c}{\includegraphics[width=0.372\linewidth]{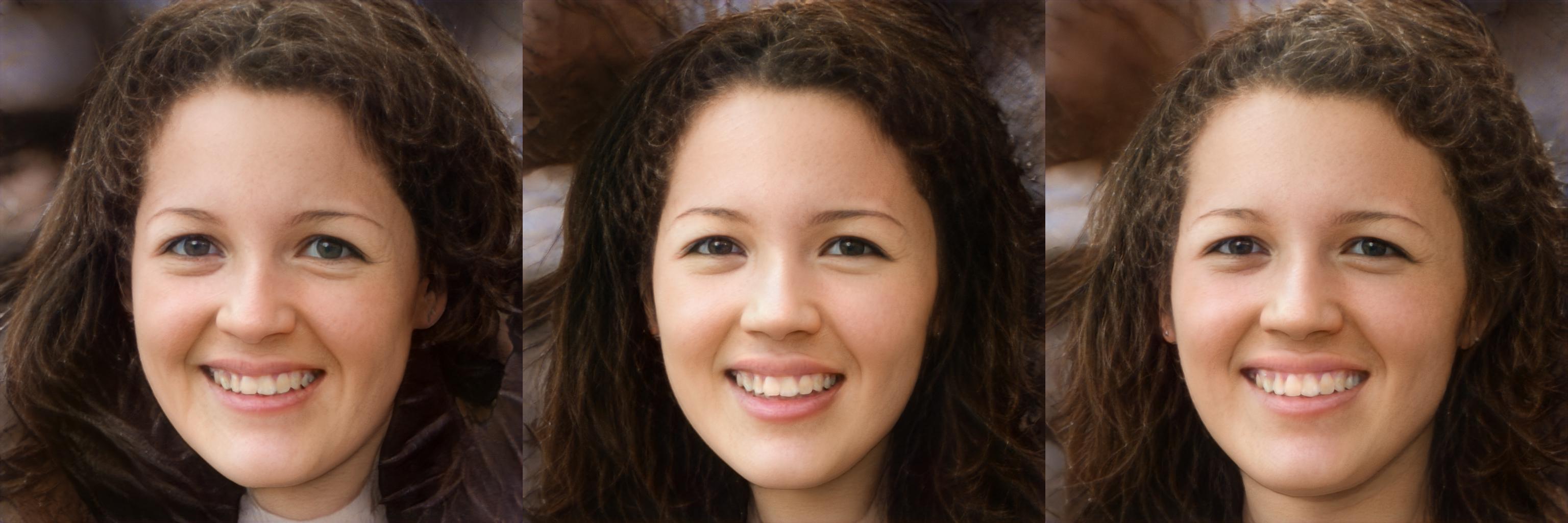}}
    &\multicolumn{3}{c}{\includegraphics[width=0.372\linewidth]{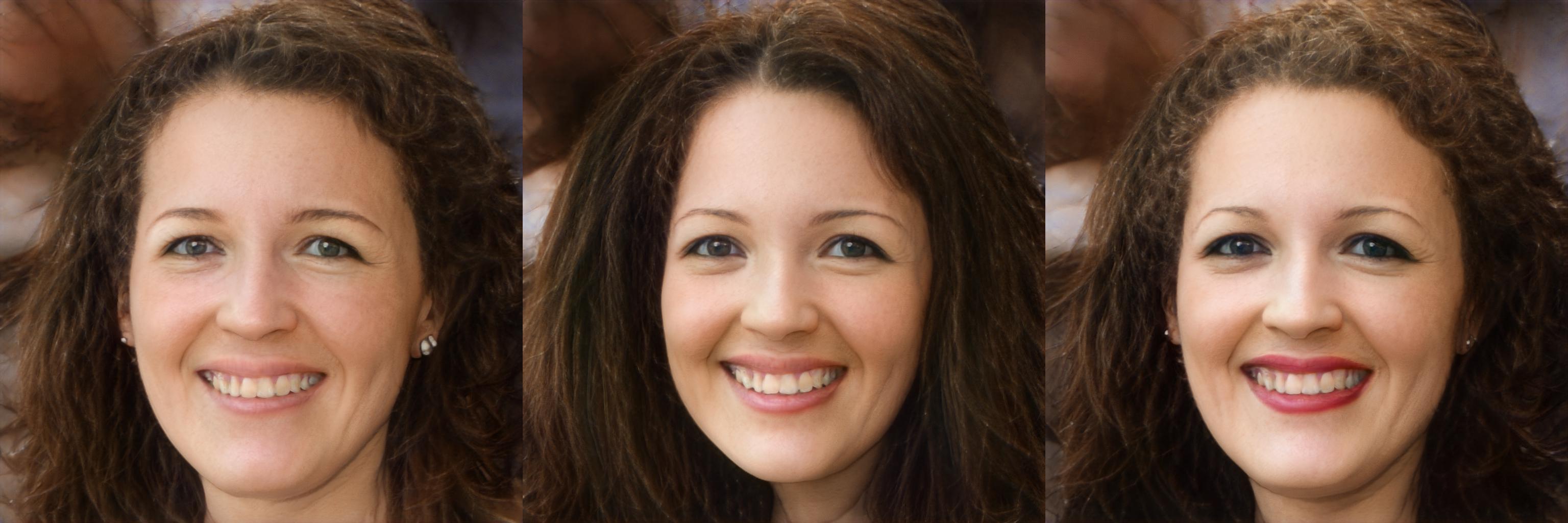}}
    \\
    Original&Projected&GANSpace&InterFaceGAN&Ours&GANSpace&InterFaceGAN&Ours
    \\
\end{tabular}
\caption{\textbf{Disentangled facial attribute editing on real images.} The first two columns show the original image and the projected image reconstructed with the encoded latent code in StyleGAN. From the 3rd column in each subfigure, from left to right are the manipulation result of GANSpace \cite{harkonen2020ganspace}, that of InterFaceGAN \cite{shen2019interpreting} and ours. Compared to recent approaches, our method achieves a controllable, disentangled and realistic editing, where the person's identity is preserved.
}
\label{comparison}
\end{figure*}
\noindent \textbf{Seamless cloning.} We use Poisson image editing method \cite{perez2003poisson} to blend the modified faces with the original input frames. In order to blend only the face area, we use the segmentation mask obtained from the detected facial landmarks.

\subsection{Implementation details}
\label{sec:implementation}

We explore disentangled manipulations in the latent space of StyleGAN2 \cite{karras2019analyzing} pretrained on FFHQ dataset \cite{karras2019style}. 
In this paper, we conduct all the experiments with the latest StyleGAN2. For simplicity, when we mention StyleGAN, it refers to the latest version - StyleGAN2.
To prepare the training data, we project images of CelebA-HQ \cite{karras2018progressive} to the latent space \set{W^+} of StyleGAN using a pre-trained encoder \cite{richardson2020encoding} and obtain the corresponding latent codes. CelebA-HQ contains $30K$ face images at $1024^2$ resolution, each annotated by $40$ facial attributes. We train a separate latent transformer for each facial attribute.  

To predict the attributes on the latent codes, we train a latent classifier on the `latent code - label' pairs, which is fixed during the training of the latent transformer.  The model is designed to predict all the $40$ attributes together, and trained with binary cross entropy loss.

For the training of the latent transformer, we use $90\%$ of the prepared data for training set and train the model for $100K$ iterations, with a batch size of $32$. The weights balancing each loss are set to $\lambda_{\rm attr}= 1$ and $\lambda_{\rm rec}=10$. We use Adam optimizer \cite{kingma14} with a learning rate of $0.001$, $\beta_1 = 0.9$ and $\beta_2 = 0.999$.

\section{Experiments}

\begin{figure*}[!ht]
\centering
\includegraphics[width=0.99\linewidth]{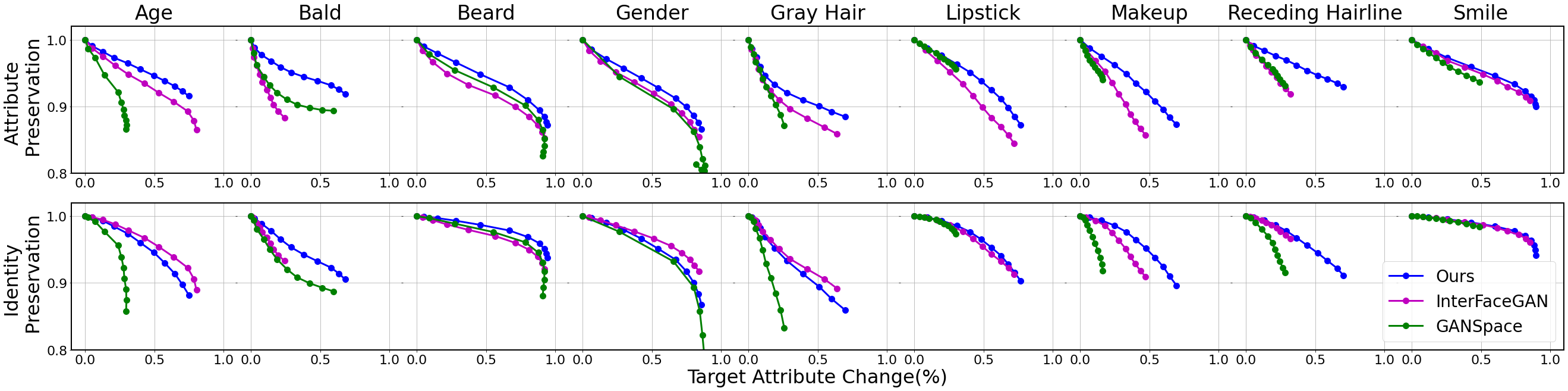}
\\
\caption{\textbf{Attribute and identity preservation vs. target attribute change.} For each method, we edit each target attribute with $10$ different scaling factors ($\{0.2\cdot d, 0.4\cdot d, ..., 2\cdot d\}$, $d$ is the %
magnitude of change suggested in each method), and generate the modified images. Attribute preservation rate and identity preservation score are measured on the output images. 
In the figure, each point corresponds to a scaling factor, where the position $x$ indicates the target attribute change rate (the fraction of the samples with target attribute successfully changed among all the manipulations). In the upper sub-figure, the position $y$ indicates the average attribute preservation rate on the other attributes. In the bottom sub-figure, the position $y$ indicates the average identity preservation score.
Ideally, we want higher attribute and identity preservation for the same amount of change on the target attribute (higher curve is better).
}
\label{eval_attr}
\end{figure*}
\begin{figure*}[t]
\centering
\small
\setlength{\tabcolsep}{0.3pt}
\renewcommand{\arraystretch}{0.5}
\begin{tabular}{P{0.14\linewidth}P{0.14\linewidth}P{0.14\linewidth}P{0.14\linewidth}P{0.14\linewidth}P{0.14\linewidth}P{0.14\linewidth}}
\centering
Original&Projected& - Chubby & + Blond Hair & + Smile & + Lipstick & + Eyeglasses
\\
\multicolumn{7}{c}{\includegraphics[width=0.99\linewidth]{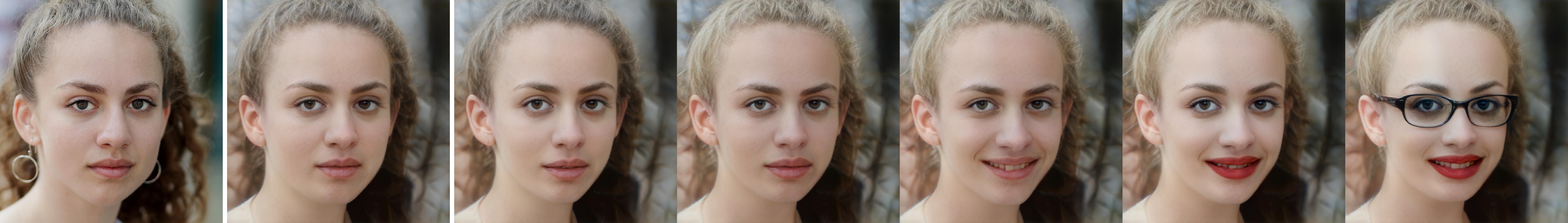}}
\\
&& - Eyeglasses & + Bangs & + \scriptsize{Bags Under Eyes}  & - Smiling & + Age
\\
\multicolumn{7}{c}{\includegraphics[width=0.99\linewidth]{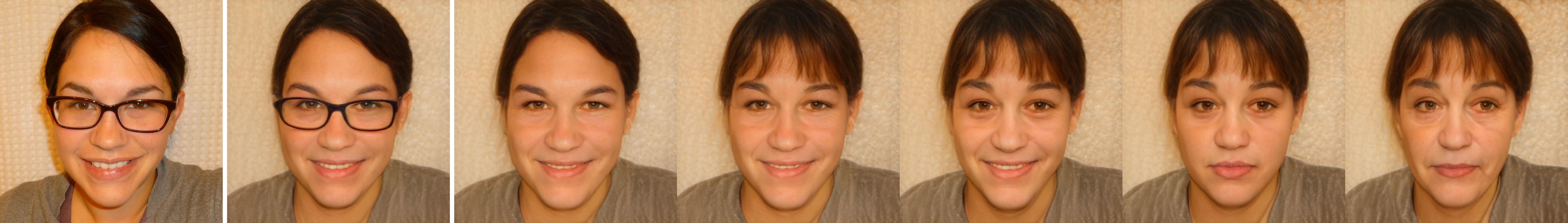}}
\\
&&  + Smiling & + Beard & + \scriptsize{Receding Hairline} & + Eyeglasses & + \scriptsize{Arched Eyebrows} 
\\
\multicolumn{7}{c}{\includegraphics[width=0.99\linewidth]{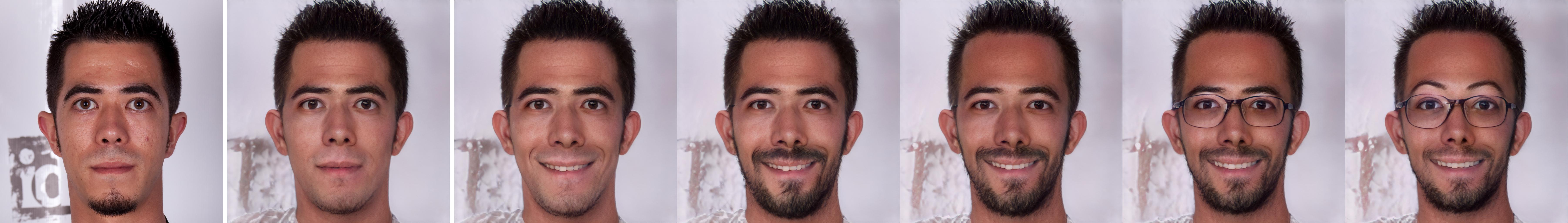}}
\\
&&  - Smiling & - Chubby & + Goatee & + Eyeglasses & + Pale Skin
\\
\multicolumn{7}{c}{\includegraphics[width=0.99\linewidth]{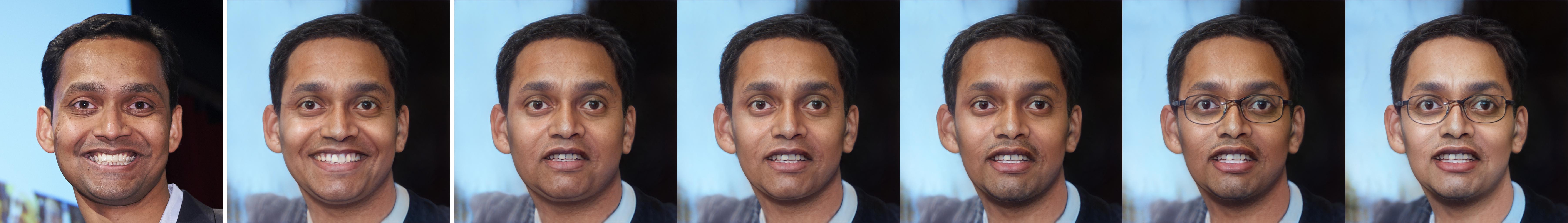}}
\\
\end{tabular} 
\caption{\textbf{Sequential facial attribute editing on real images}. Given an input image, we manipulate a list of attributes sequentially, where each time a single attribute is modified from the previous latent representation. %
}
\label{sequential}
\end{figure*}

\subsection{Disentangled manipulation on real images}

We compare our results with two state-of-the-art methods: InterFaceGAN \cite{shen2019interpreting} and GANSpace \cite{harkonen2020ganspace}. 
For a fair comparison, we follow the methodology of InterFaceGAN and train their model on StyleGAN2 for the attributes of CelebA-HQ using their official code. 
The official implementation of GANSpace on StyleGAN2 is available. For the evaluation data we use FFHQ, independent from the training of all methods. We project the real images of FFHQ to the latent space \set{W^+} of StyleGAN using the pre-trained encoder \cite{richardson2020encoding}, and manipulate the latent codes using each method with the suggested magnitude of edits ($3$ for InterFaceGAN, specified range based on attributes for GANSpace and $1$ for our method). Figure \ref{comparison} compares the manipulation results on the attributes which are available for all methods (`gender', `age', `beard' and `makeup'). Our method achieves better disentangled manipulations. For example, when changing `gender', both GANSpace and InterFaceGAN modify the hairstyle, and when changing `age', GANSpace adds eyeglasses and InterFaceGAN affects smile. In contrast, our method succeeds to separate hairstyle from `gender' and disentangle `eyeglasses' from `age', thanks to the attribute and latent code regularization terms.
The directions of GANSpace are discovered from PCA so that they may control several attributes simultaneously. For InterFaceGAN, no attribute preservation is applied when searching the semantic boundary. Compared with their methods, our editing results are of better visual quality and preserve the original facial identities better.

\subsection{Quantitative evaluation}
We compare our method quantitatively with GANSpace and InterFaceGAN using three metrics: target attribute change rate, attribute preservation rate and identity preservation score. Given a set of manipulated samples, the target attribute change rate refers to the percentage of the samples with target attribute varied among all the samples. The attribute preservation rate indicates the proportion of unchanged samples on the other attributes apart from the target. The identity preservation score refers to the average cosine similarity between the VGG-Face \cite{parkhi2015deep} embeddings of the original projected images and the manipulated results. 

For the evaluation data, we project the first $1K$ images of FFHQ into the the latent space \set{W^+} of StyleGAN using the pre-trained encoder \cite{richardson2020encoding}. For each input image and each method, we edit each attribute with $10$ different scaling factors ($\{0.2\cdot d, 0.4\cdot d, ..., 2\cdot d\}$, $d$ is the magnitude of change suggested by each method) and generate the corresponding images. To predict the attributes on the modified images, we use a state-of-the-art facial attribute classifier \cite{he2018harnessing}, independent from all methods. Based on the classification result, we consider an attribute active if its probability is greater than $0.5$, otherwise inactive (\ie, w/ bangs versus w/o bangs). For each scaling factor, we compute the target attribute change rate, and the attribute preservation rate averaged on the other attributes. To check the identity preservation, we compute the average identity preservation score. 
Figure \ref{eval_attr} presents the attribute and identity preservation w.r.t. the target attribute change on the attributes detected by all methods. For attributes like `beard', `gender' and `smile', all the methods handle well. For other attributes, we observe that for the same amount of change on the target attribute, our approach has a higher attribute preservation rate while achieving a comparable or better identity preservation score. Overall our method achieves better disentanglement and better identity preservation than existing methods.

\subsection{Sequential editing} 

Thanks to the disentanglement property of our approach, it achieves sequential modifications of several attributes on real images. 
We project real images of FFHQ to the latent space \set{W^+} of StyleGAN using the pre-trained encoder \cite{richardson2020encoding}, and apply manipulations on a list of attributes sequentially. As shown in Figure \ref{sequential}, our method achieves disentangled and realistic modifications, and is not limited to a defined order of attributes.

\begin{figure}
\centering
\small
\setlength{\tabcolsep}{0pt}
\renewcommand{\arraystretch}{0.5}
\begin{tabular}{P{0.04\linewidth}P{0.24\linewidth}P{0.24\linewidth}P{0.24\linewidth}P{0.24\linewidth}}
& Original & $\lambda_{\rm attr}=0$ & $\lambda_{\rm rec}=0$ & Ours 
\\
\rotatebox{90}{\quad Gender} 
&\includegraphics[width=\linewidth]{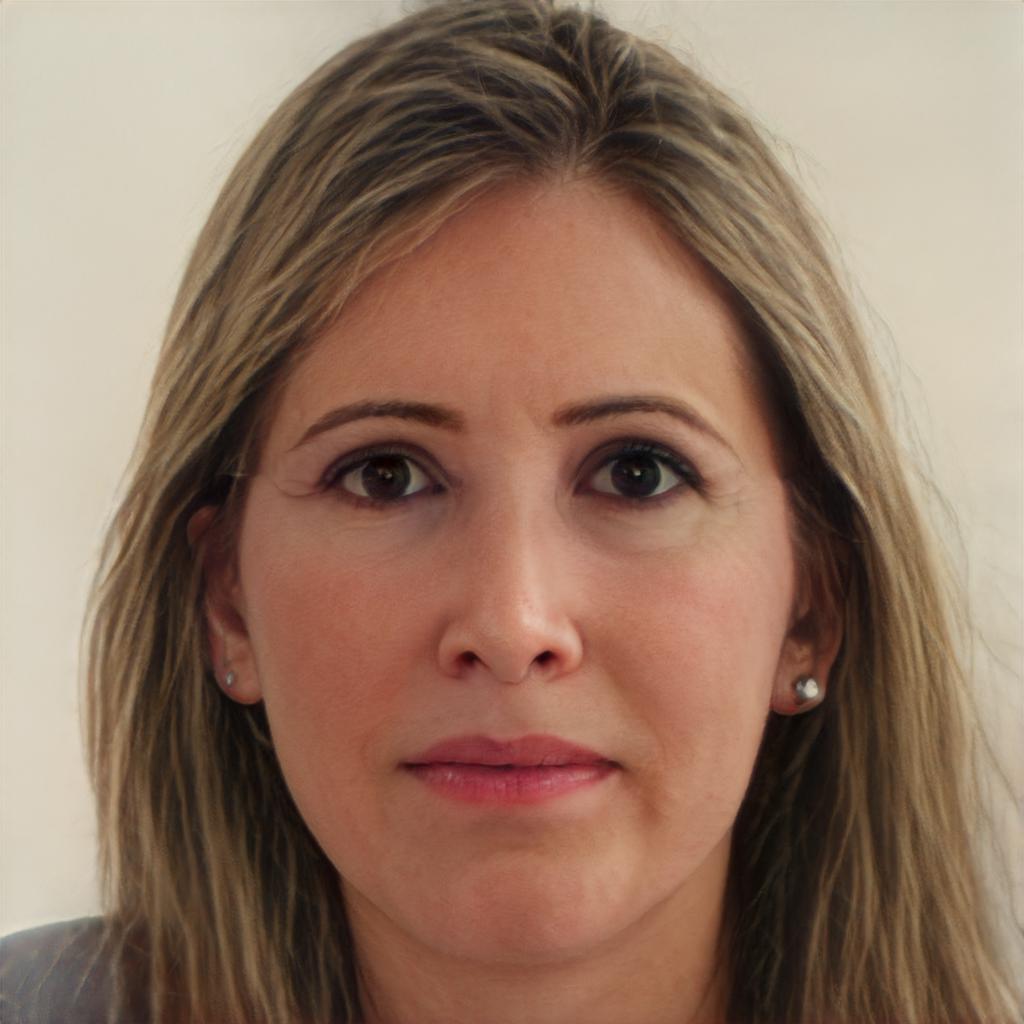}
&\multicolumn{3}{c}{\includegraphics[width=0.72\linewidth]{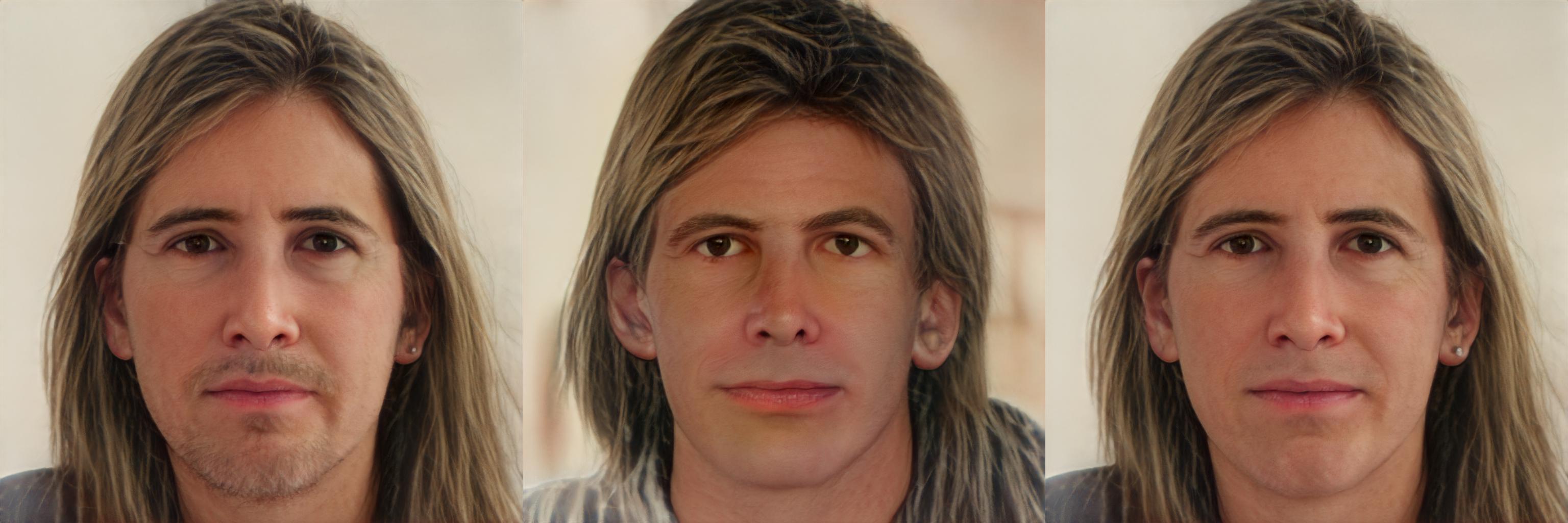}}
\\
\rotatebox{90}{\quad Eyeglasse} 
&\includegraphics[width=\linewidth]{fig/ablation/00059.jpg}
&\multicolumn{3}{c}{\includegraphics[width=0.72\linewidth]{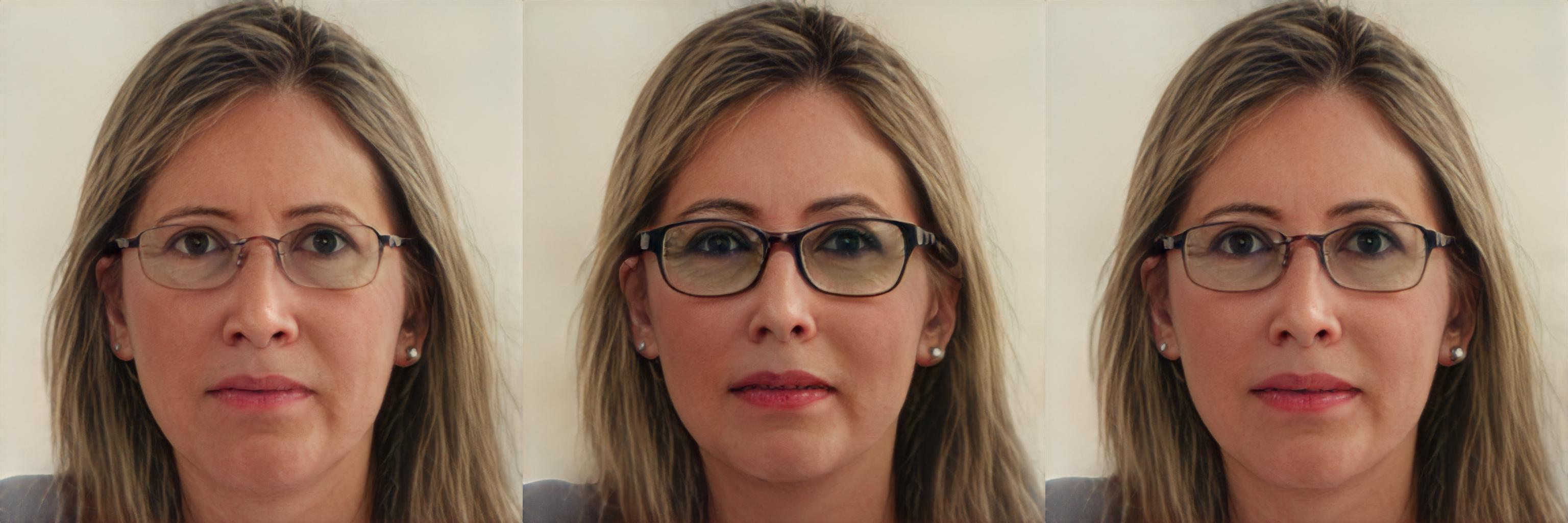}}
\\
\rotatebox{90}{\quad Young} 
&\includegraphics[width=\linewidth]{fig/ablation/00059.jpg}
&\multicolumn{3}{c}{\includegraphics[width=0.72\linewidth]{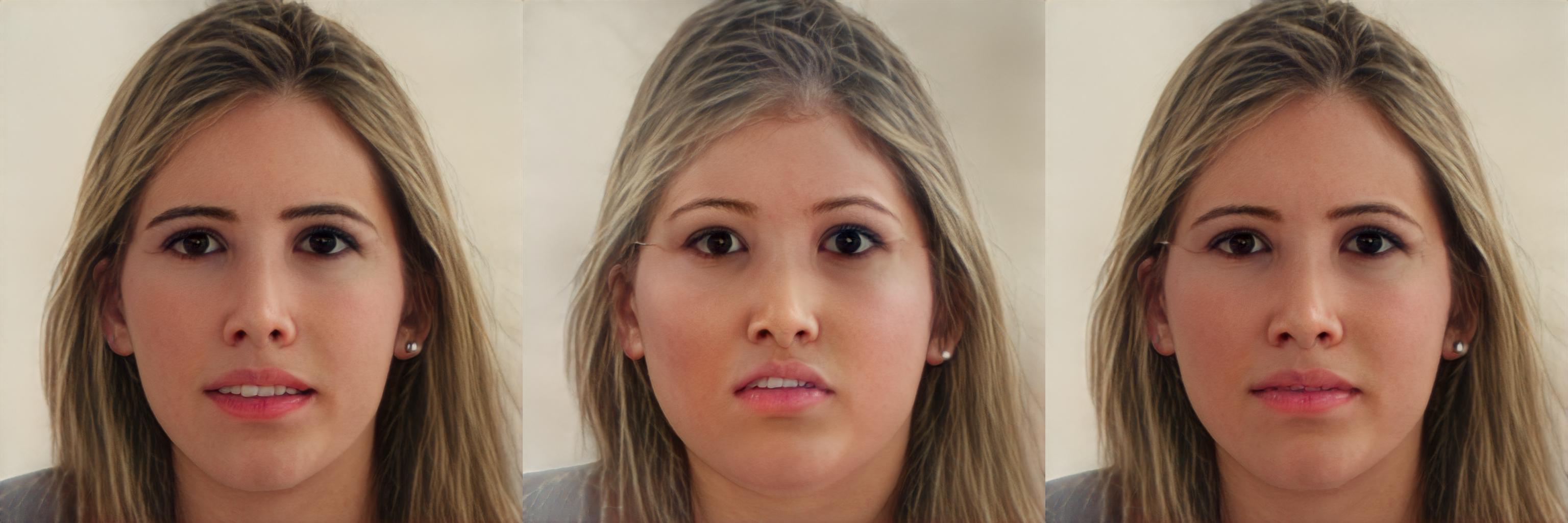}}
\\
\end{tabular}
\caption{Ablation study on the loss composition in Eq. (\ref{eq:total}). Each row corresponds to a single attribute editing. From left to right: original image, manipulation results of different scenarios. When $\lambda_{\rm attr}=0$, editing target attribute affects the others. When $\lambda_{\rm rec}=0$ it fails to preserve the facial identity. Our proposed baseline yields better disentangled results with identity preserved.
}
\label{loss_compare}
\end{figure}

\begin{figure}[t]
\centering
\includegraphics[width=0.99\linewidth]{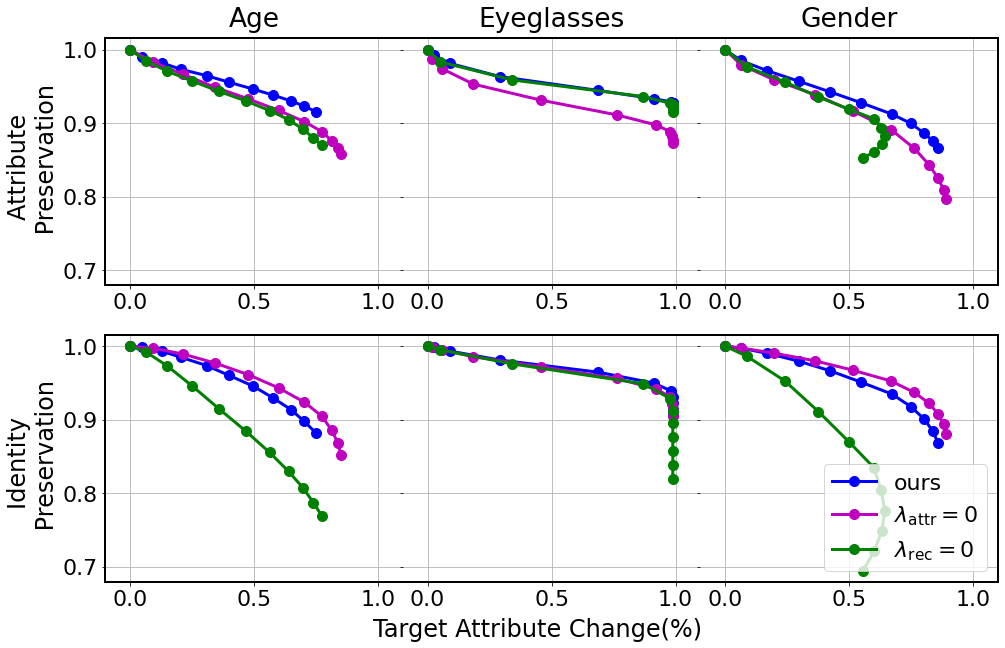}
\caption{Quantitative comparison of different loss compositions. For each scenario, we edit each attribute with $10$ scaling factors ($\{0.2, 0.4, ..., 2\}$) and measure the attribute preservation rate and identity preservation score on the modified images. Each point marker represents a scaling factor. The upper sub-figure presents the average attribute preservation rate w.r.t. target attribute change rate. The bottom sub-figure presents the average identity preservation score w.r.t. target attribute change rate. Our chosen baseline has a better trade-off between attribute and identity preservation.}
\label{ablation_loss}
\end{figure}
\begin{figure*}[t]
\centering
\small
\setlength{\tabcolsep}{1pt}
\renewcommand{\arraystretch}{0.5}
\begin{tabular}{c c}
\rotatebox{90}{\quad Input frames} &
\includegraphics[width=0.98\linewidth]{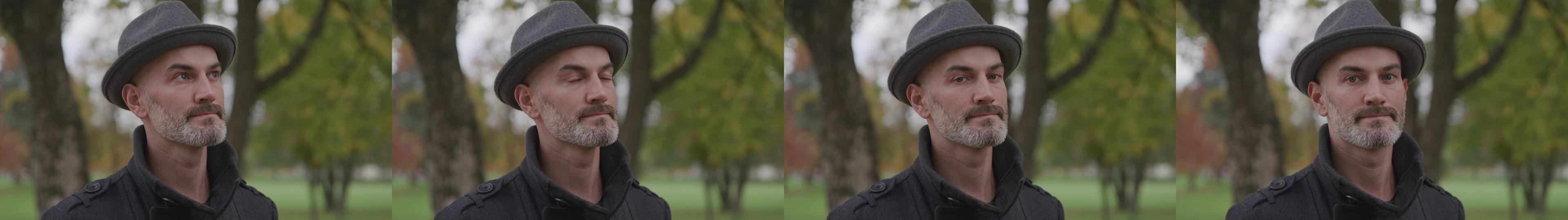}
\\
\rotatebox{90}{\quad \quad - Beard} &
\includegraphics[width=0.98\linewidth]{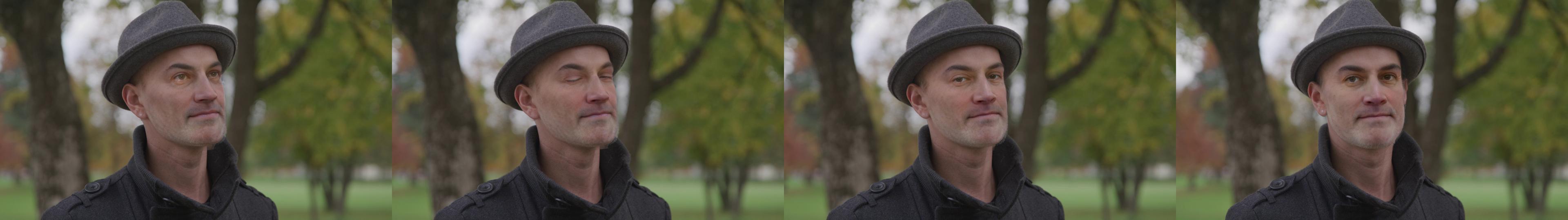}
\\
\rotatebox{90}{\quad Input frames} &
\includegraphics[width=0.98\linewidth]{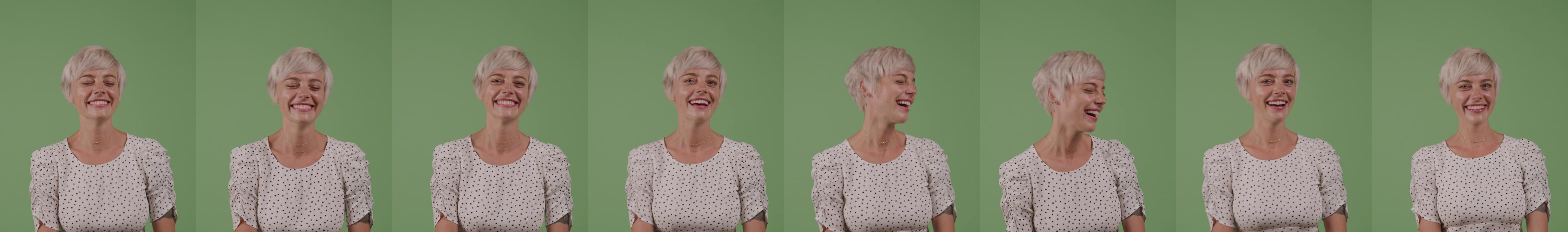}
\\
\rotatebox{90}{\quad + Makeup} &
\includegraphics[width=0.98\linewidth]{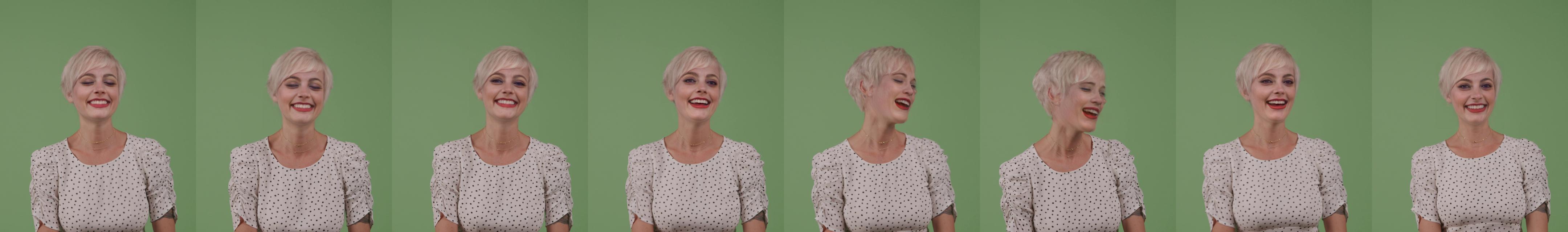}
\end{tabular}
\caption{\textbf{Facial attribute editing on videos.} In each sub-figure, the top row shows the input frames, the bottom row shows the output frames obtained from our proposed video manipulation pipeline. A face image is cropped and aligned from each frame, and encoded to latent space of StyleGAN. The encoded latent code is passed into the latent transformer to get the target attribute varied, then decoded to an output face and blended with the input frame.
}
\label{video_compare}
\end{figure*}
\subsection{Loss analysis} 

We carry out an ablation study to analyze the effects of each regularization terms in Eq. (\ref{eq:total}). In our proposed baseline, the weights balancing each regularization term are set to $\lambda_{\rm attr}=1$ and $\lambda_{\rm rec}=10$. We compare with two different scenarios: $\lambda_{\rm attr}=0$ (w/o attribute regularization) and $\lambda_{\rm rec}=0$ (w/o latent regularization). As shown in Figure \ref{loss_compare}, when $\lambda_{\rm attr}=0$ the output is not only manipulated on the target attribute but also affected on the other attributes, \eg, beard added when changing gender, mouth affected when changing age. When $\lambda_{\rm rec}=0$, the manipulated images fail to preserve the original facial identity. Balancing each term, our proposed baseline achieves attribute editing with better disentanglement and identity preservation. Figure \ref{ablation_loss} provides a quantitative comparison of the three scenarios. As can be observed, our chosen baseline preserves the other attributes best, with the same amount of attribute change, without sacrificing the identity preservation.

\subsection{Video editing}

We apply our manipulation method on real-world videos collected from FILMPAC library \cite{filmpac}. Figure \ref{video_compare} shows the qualitative results of facial attribute editing on videos obtained from our proposed pipeline. From each input frame, we crop and align a face image and encode it to the latent space of StyleGAN with a pre-trained encoder \cite{richardson2020encoding}.
The encoded latent code is processed by the latent transformer to vary the target attribute, then decoded to an output face image and blended with the input frame. As can be seen from the results, our proposed method succeeds in removing the facial hair or adding the makeup, without influencing the consistency between the frames. Nevertheless, we also observe that the proposed method has more difficulty handling extreme pose (side face), which may be due to the limitation of the generation capacity of the StyleGAN model. Please refer to the appendix to check the videos and see more video results.


\section{Conclusion and future work}

In this paper, we have presented a latent transformation network to perform facial attribute editing in real images and videos via the latent space of StyleGAN. Our method generates realistic manipulations with better disentanglement and identity preservation than other approaches. We have extended our method to the case of videos, achieving stable and consistent modifications. To the best of our knowledge, this is the first work to present stable facial attribute editing on high resolution videos. 
Some future work could be dedicated to improve both the applicability of the method and the performance on videos. In particular, the method has difficulty handling side poses due to the fact that StyleGAN has difficulties in generating faces in side poses. This could be potentially addressed by jointly training the StyleGAN encoder with the generator, or training an improved StyleGAN generator using images with better diversity of poses.

{\small
\bibliographystyle{ieee_fullname}
\bibliography{egbib}
}

\clearpage
\appendix
\section*{Appendix}

We provide further details on the experiments described in the main paper. Sec. \ref{sup:image_manip} presents additional results of disentangled attribute manipulation and sequential attribute manipulation on real images of FFHQ dataset \cite{karras2019style}. Sec. \ref{sup:video_manip} provides additional results of facial attribute editing on videos collected from FILMPAC library \cite{filmpac} and RAVDESS dataset \cite{livingstone2018ryerson}. 
We show that our method handles disentangled and sequential facial attribute manipulation on images and videos. Please refer to our project page to view the facial attribute editing on videos: \url{https://github.com/InterDigitalInc/latent-transformer/tree/master/image}.

\section{More results on real image editing}
\label{sup:image_manip}

\begin{figure*}[t]
\centering
\textit{\large{Single attribute manipulation}}
\\
\small
\setlength{\tabcolsep}{0.3pt}
\renewcommand{\arraystretch}{0.5}
\begin{tabular}{P{0.135\linewidth}P{0.135\linewidth}P{0.135\linewidth}P{0.135\linewidth}P{0.135\linewidth}P{0.135\linewidth}P{0.135\linewidth}}
\centering
Original&Projected& \scriptsize{Arched Eyebrows} & - Chubby & - Pointy Nose & - Smiling & Pale Skin
\\
\multicolumn{7}{c}{\includegraphics[width=0.94\linewidth]{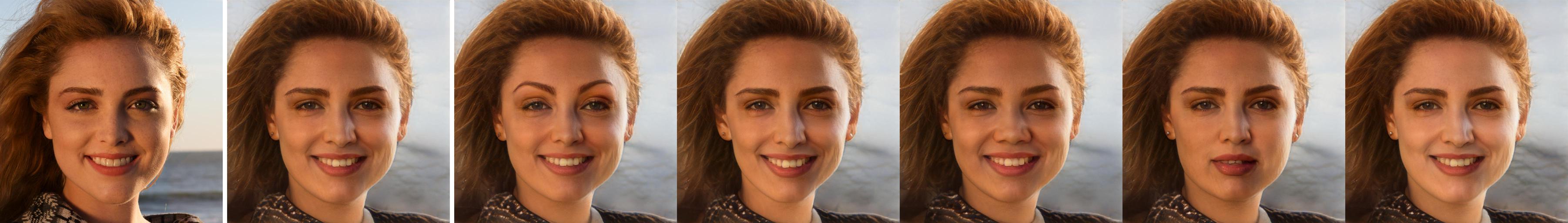}}
\\
Original&Projected& Black Hair & Smiling & \scriptsize{Bags Under Eyes}  & - Age & Eyeglasses
\\
\multicolumn{7}{c}{\includegraphics[width=0.94\linewidth]{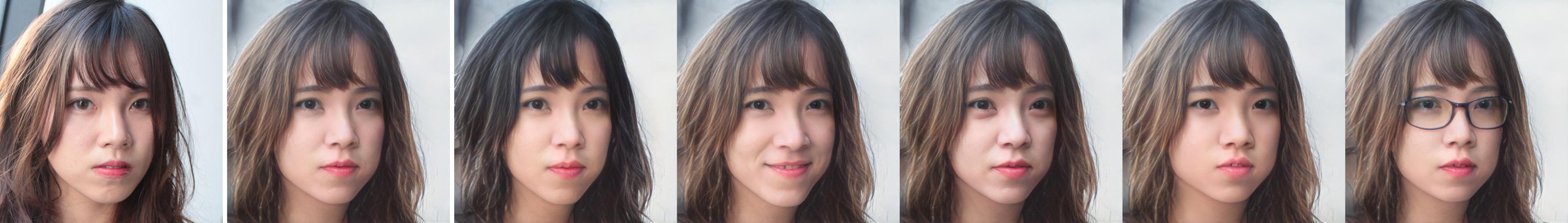}}
\\
Original&Projected&  - Chubby & Bangs & Smiling & Age & Makeup
\\
\multicolumn{7}{c}{\includegraphics[width=0.94\linewidth]{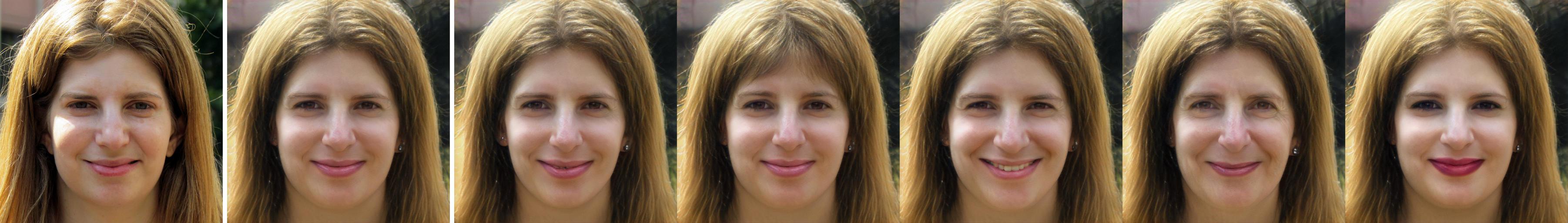}}
\\
Original&Projected&  - Chubby & - Smiling & - Narrow Eyes & Makeup & Eyeglasses
\\
\multicolumn{7}{c}{\includegraphics[width=0.94\linewidth]{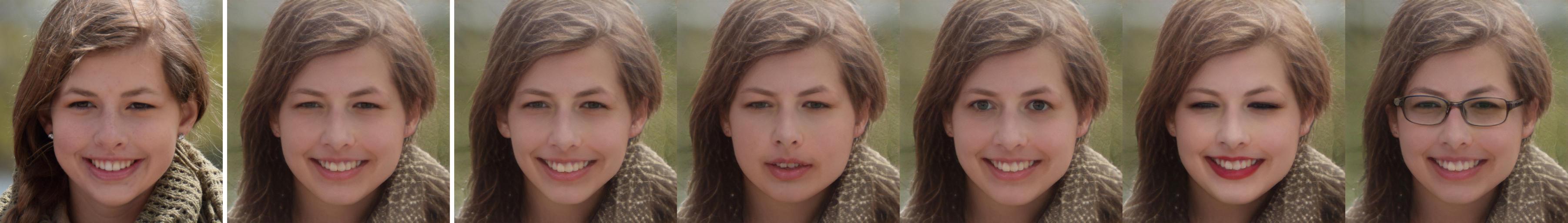}}
\\
Original&Projected&  - Eyeglasses & Smiling & Goatee & \scriptsize{Arched Eyebrows} & - Age
\\
\multicolumn{7}{c}{\includegraphics[width=0.94\linewidth]{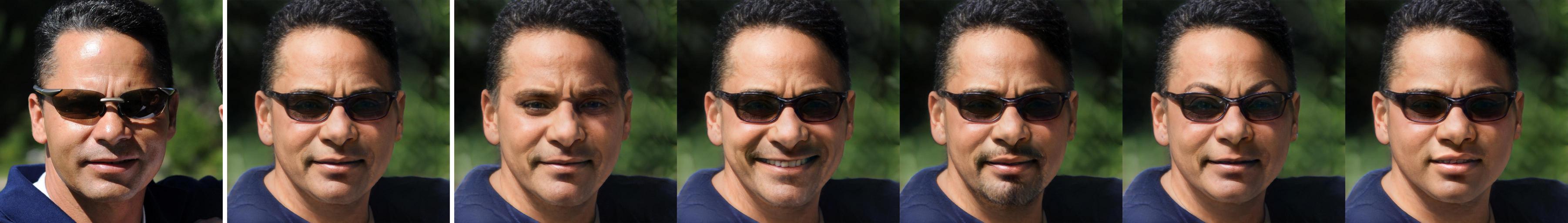}}
\\
Original&Projected&  Beard & \scriptsize{Bushy Eyebrows} & \scriptsize{Mouth Slightly Open} & \scriptsize{Receding Hairline} & Eyeglasses
\\
\multicolumn{7}{c}{\includegraphics[width=0.94\linewidth]{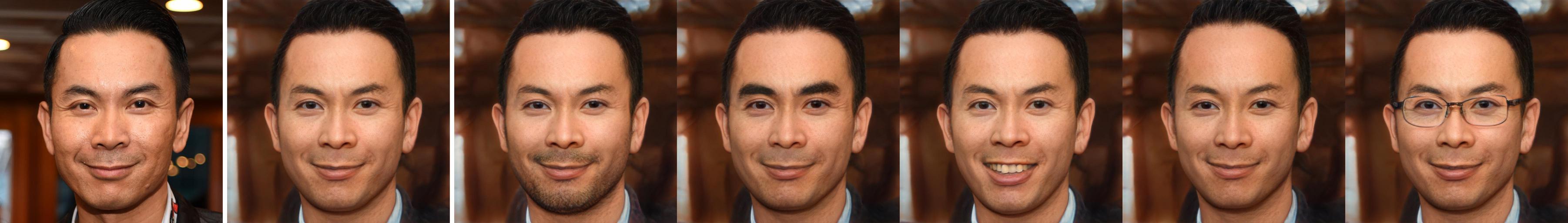}}
\\
Original&Projected&  - Smiling & - Chubby & Beard & Eyeglasses & \scriptsize{Receding Hairline}
\\
\multicolumn{7}{c}{\includegraphics[width=0.94\linewidth]{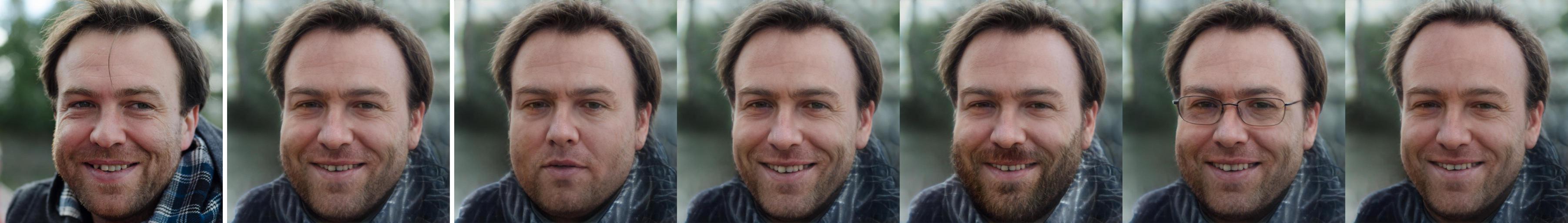}}
\\
Original&Projected&  Smiling & Goatee & - Eyeglasses & - \scriptsize{Arched Eyebrows} & - Age
\\
\multicolumn{7}{c}{\includegraphics[width=0.94\linewidth]{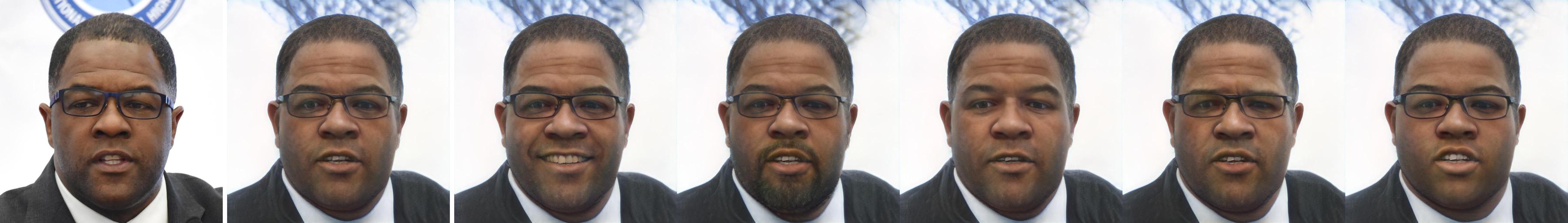}}
\\
\end{tabular} 
\caption{\textbf{Single attribute editing on real images}. Given an input image, we show manipulation results on several attributes, where each time a single attribute is modified from the first projected latent representation. %
}
\label{sup:disentangle}
\end{figure*}

\begin{figure*}[t]
\centering
\textit{\large{Sequential attribute manipulation}}
\\
\small
\setlength{\tabcolsep}{0.3pt}
\renewcommand{\arraystretch}{0.5}
\begin{tabular}{P{0.135\linewidth}P{0.135\linewidth}P{0.135\linewidth}P{0.135\linewidth}P{0.135\linewidth}P{0.135\linewidth}P{0.135\linewidth}}
\centering
Original&Projected& + \scriptsize{Arched Eyebrows} & - Chubby & - Pointy Nose & - Smiling & + Pale Skin
\\
\multicolumn{7}{c}{\includegraphics[width=0.94\linewidth]{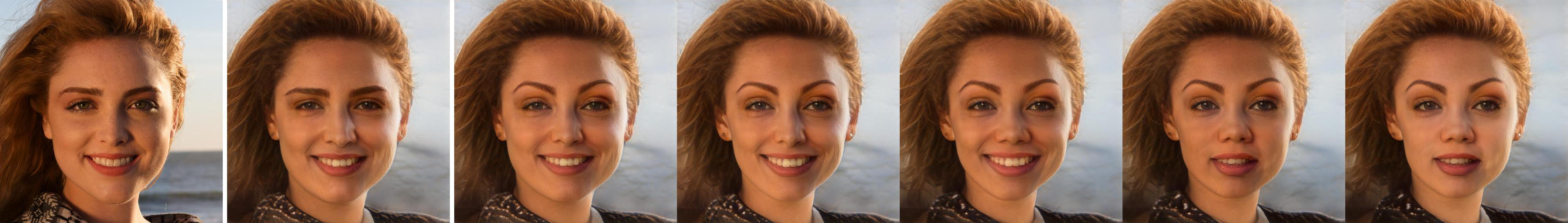}}
\\
Original&Projected& + Black Hair & + Smiling & + \scriptsize{Bags Under Eyes}  & - Age & + Eyeglasses
\\
\multicolumn{7}{c}{\includegraphics[width=0.94\linewidth]{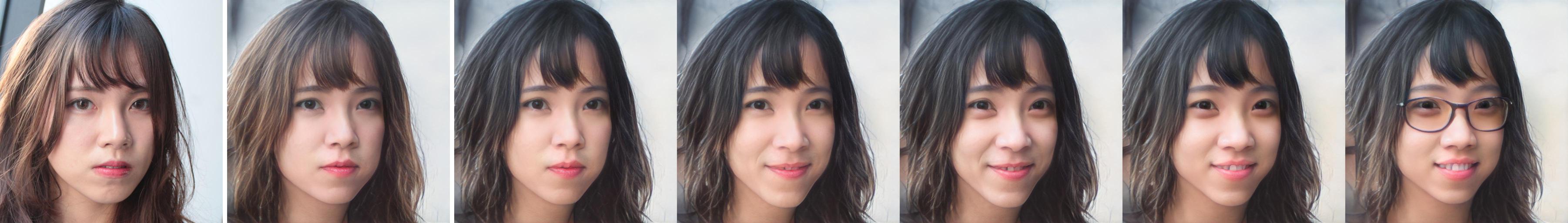}}
\\
Original&Projected&  - Chubby & + Bangs & + Smiling & + Age & + Makeup
\\
\multicolumn{7}{c}{\includegraphics[width=0.94\linewidth]{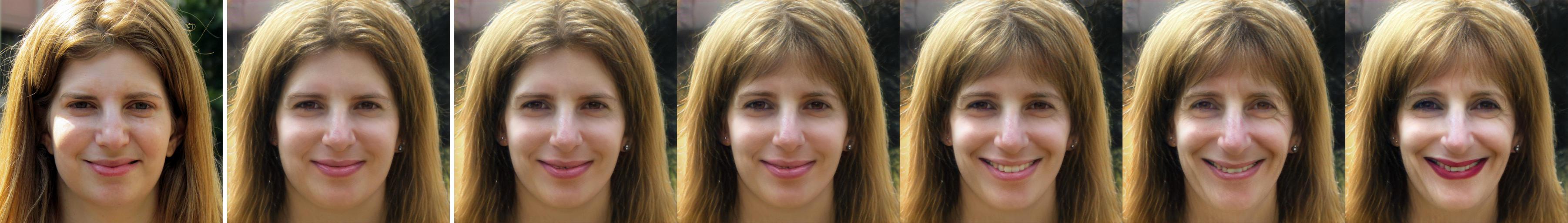}}
\\
Original&Projected&  - Chubby & - Smiling & - Narrow Eyes & + Makeup & + Eyeglasses
\\
\multicolumn{7}{c}{\includegraphics[width=0.94\linewidth]{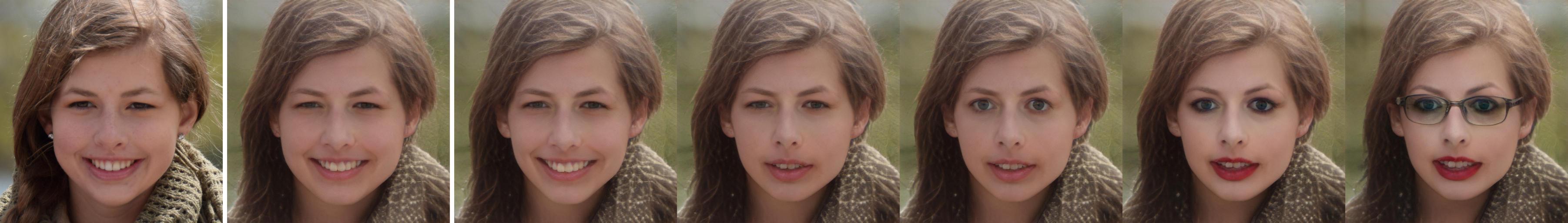}}
\\
Original&Projected&  - Eyeglasses & + Smiling & + Goatee & + \scriptsize{Arched Eyebrows} & - Age
\\
\multicolumn{7}{c}{\includegraphics[width=0.94\linewidth]{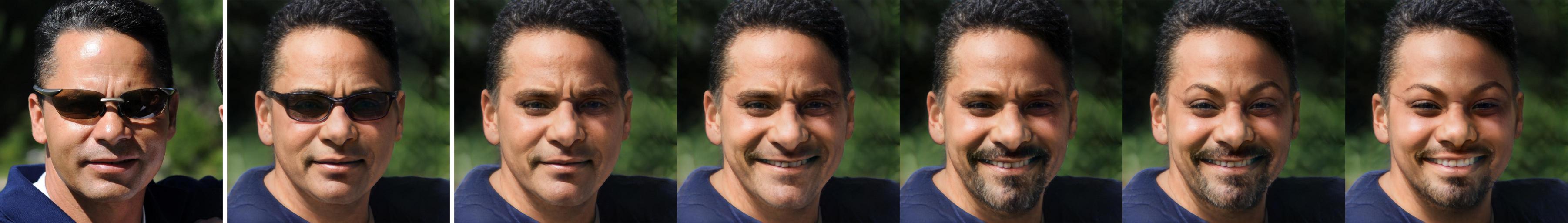}}
\\
Original&Projected&  + Beard & + \scriptsize{Bushy Eyebrows} & + \scriptsize{Mouth Slightly Open} & + \scriptsize{Receding Hairline} & + Eyeglasses
\\
\multicolumn{7}{c}{\includegraphics[width=0.94\linewidth]{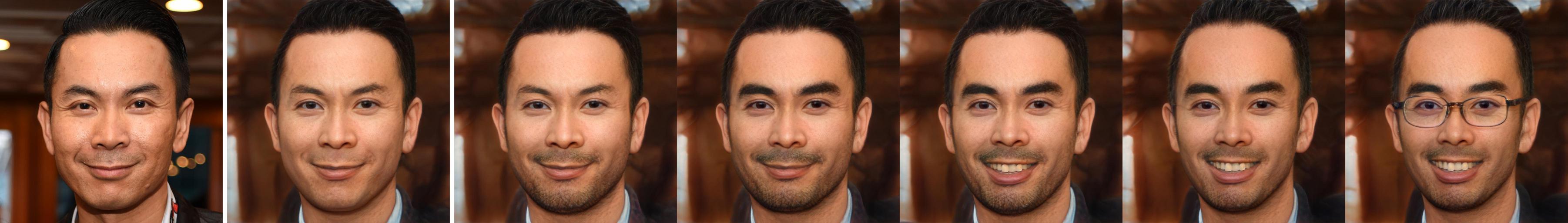}}
\\
Original&Projected&  - Smiling & - Chubby & + Beard & + Eyeglasses & + \scriptsize{Receding Hairline}
\\
\multicolumn{7}{c}{\includegraphics[width=0.94\linewidth]{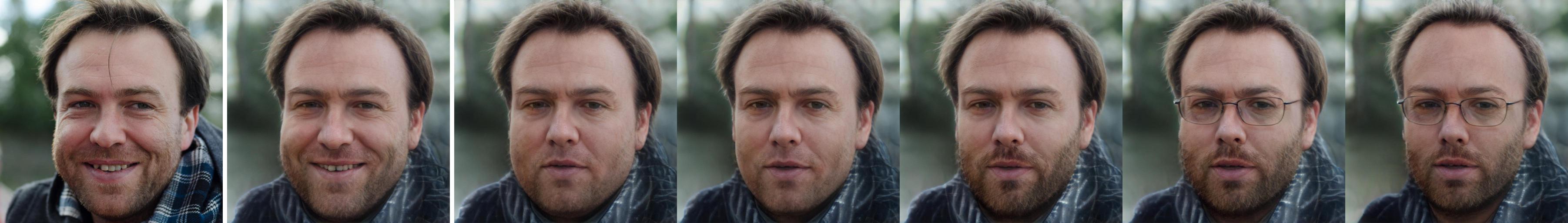}}
\\
Original&Projected&  + Smiling & + Goatee & - Eyeglasses & - \scriptsize{Arched Eyebrows} & - Age
\\
\multicolumn{7}{c}{\includegraphics[width=0.94\linewidth]{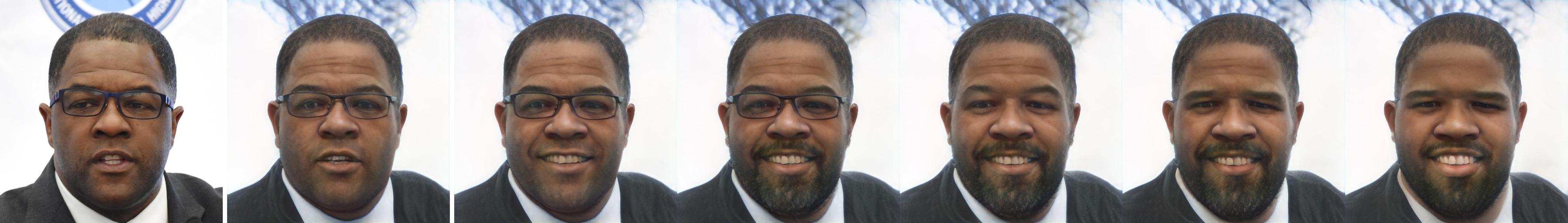}}
\\
\end{tabular} 
\caption{\textbf{Sequential attribute editing on real images}. Given an input image, we manipulate a list of attributes sequentially, where each time a single attribute is modified on top of the previous modfications. %
}
\label{sup:sequential}
\end{figure*}

\begin{figure*}[t]
\centering
\small
\setlength{\tabcolsep}{0pt}
\renewcommand{\arraystretch}{0.5}
\begin{tabular}{P{0.16\linewidth}P{0.16\linewidth}P{0.16\linewidth}}
Original & Projected & + Wavy Hair
\\
\multicolumn{3}{c}{\includegraphics[width=0.48\linewidth]{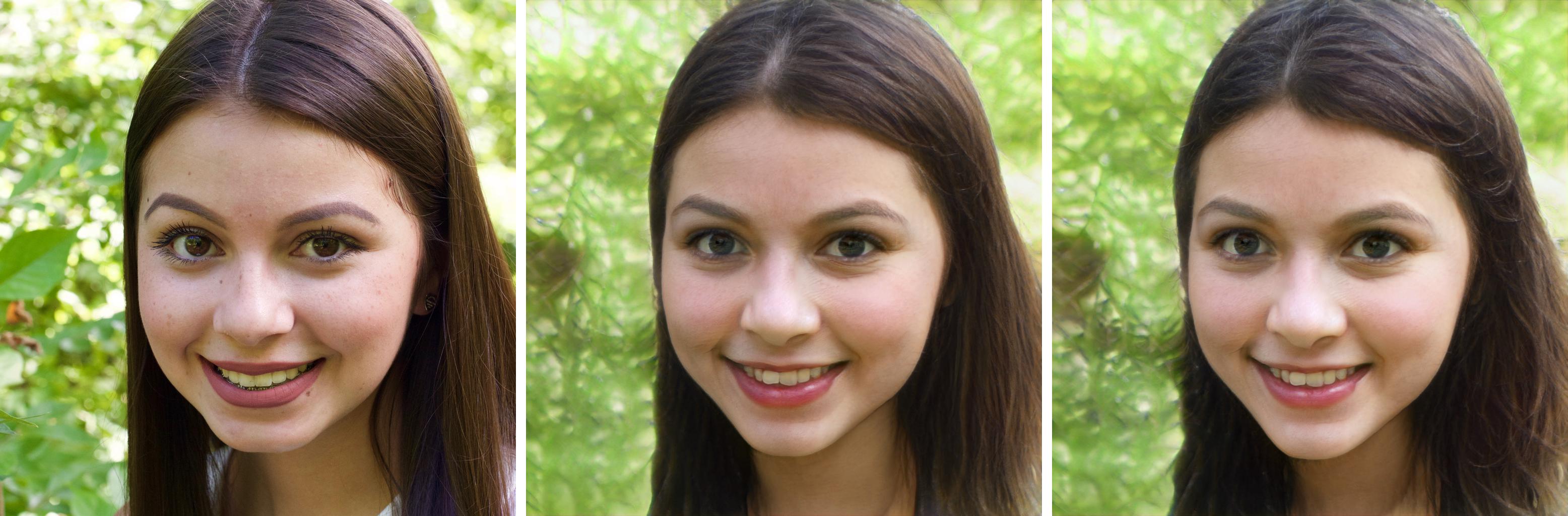}}
\\
Original & Projected & + Wearing Hat
\\
\multicolumn{3}{c}{\includegraphics[width=0.48\linewidth]{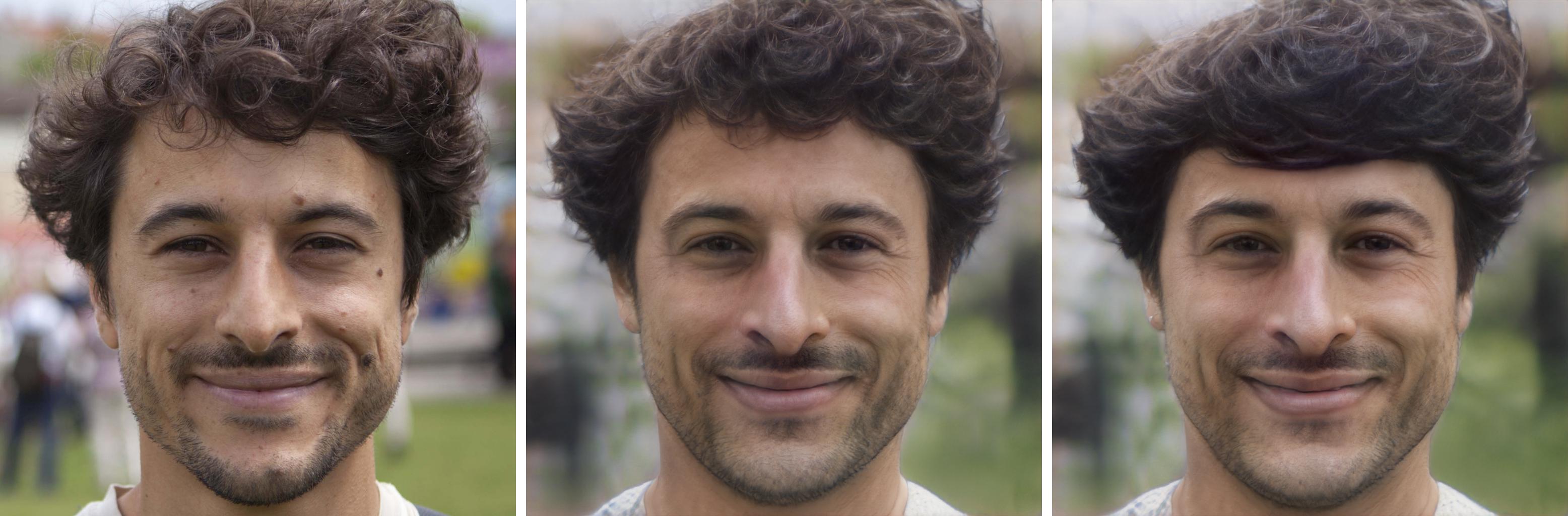}}
\\
\end{tabular}
\caption{Failure case of attribute manipulation on real images. Each row corresponds to the manipulation of an attribute. From left to right: the original image, the projected image and the manipulated image. For `wavy hair', our model yields only slight changes. In the case of `wearing hat', we fail to generate a real hat.
}
\label{sup:fail_img}
\end{figure*}

We provide additional results of disentangled attribute manipulation on real images in Figure \ref{sup:disentangle}, where only one attribute is modified at a time from the first projected image. 
Figure \ref{sup:sequential} presents additional results on sequential attribute manipulation. Here, we successively manipulate a list of attributes, meaning that each modification is performed on top of all previous modifications. 
We have trained a separate latent transformer for each of the $40$ attributes in CelebA-HQ dataset \cite{karras2018progressive}. Our method generates disentangled and identity-preserving manipulations for most of the attributes. We show some failure cases in Figure \ref{sup:fail_img}. When changing `wavy hair', only slight changes appear in the hair. One possible reason is that the hair structures are controlled by the noise inputs in StyleGAN \cite{karras2019analyzing}, while the pre-trained encoder \cite{richardson2020encoding} uses fixed noise inputs during training, which is a reasonable choice as the noise inputs have too many degrees of freedom to be reconstructed. In the case of `wearing hat', we fail to generate a real hat. This attribute is very unbalanced in CelebA-HQ, so that it is difficult to learn the correct transformation. 

\section{More results on video manipulation}
\label{sup:video_manip}

\begin{figure*}[t]
\centering
\small
\setlength{\tabcolsep}{1pt}
\renewcommand{\arraystretch}{1}
\begin{tabular}{P{0.49\linewidth}P{0.49\linewidth}}
\\
\multicolumn{1}{l}{(a) 1\_man\_with\_hat\_Arched\_Eyebrows\_Beard.avi}
&\multicolumn{1}{l}{(b) 2\_woman\_with\_bricks\_Eyeglasses\_Age.avi}
\\
\multicolumn{1}{l}{\quad - Arched Eyebrows, - Beard}
&\multicolumn{1}{l}{\quad + Eyeglasses, + Age}
\\
\includegraphics[width=\linewidth]{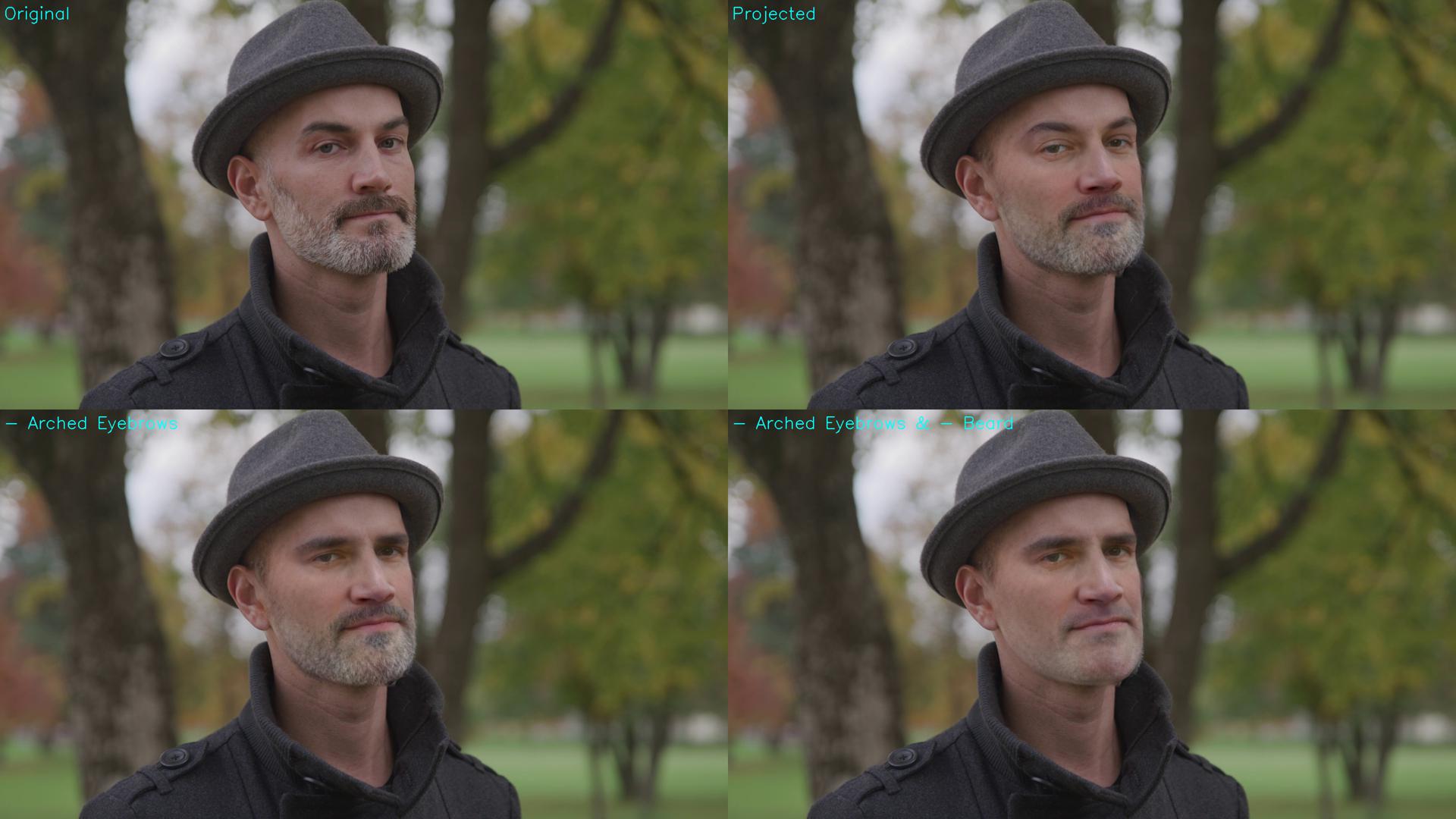}
&\includegraphics[width=\linewidth]{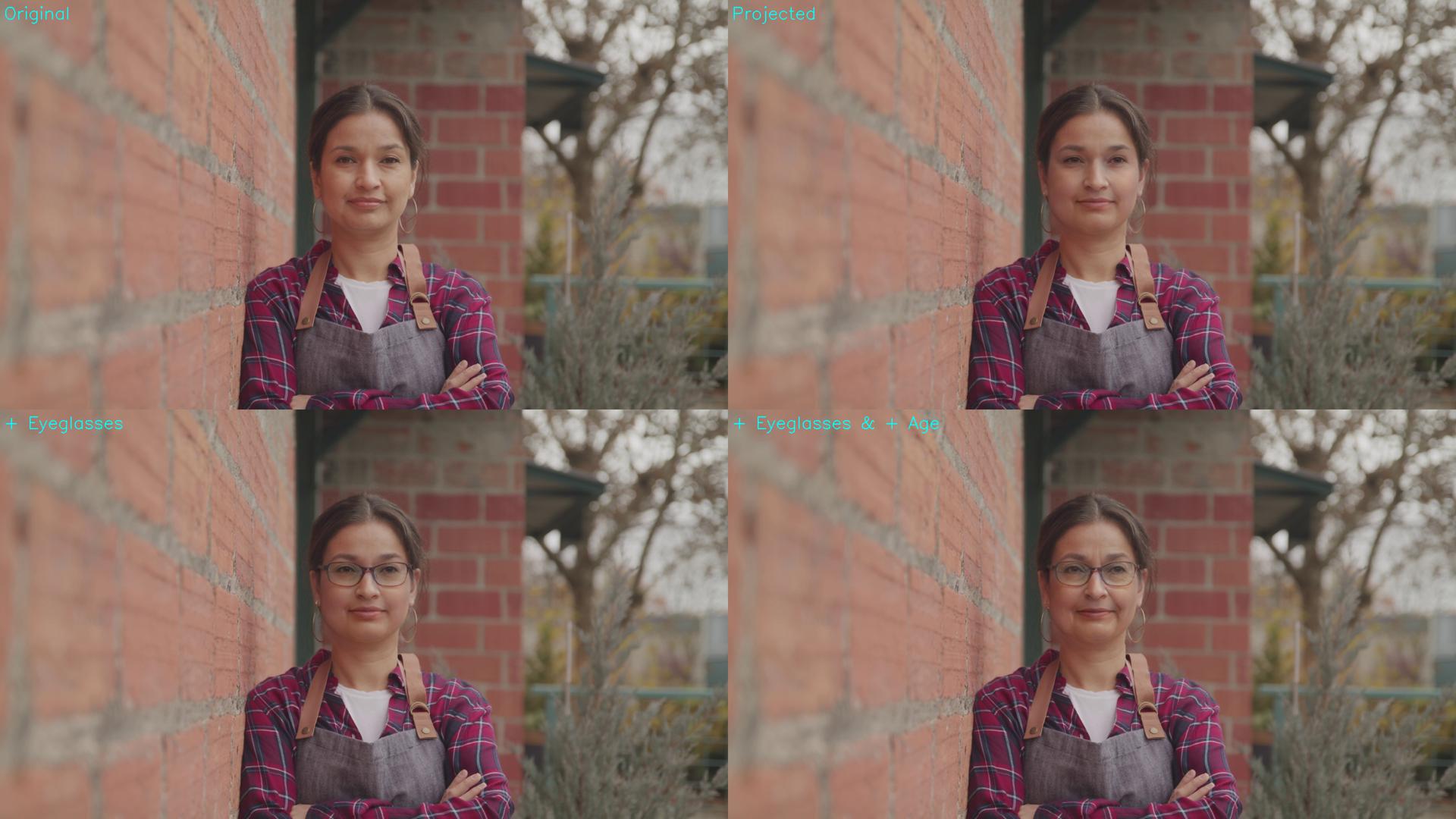}
\\
\hline
\\
\multicolumn{1}{l}{(c) 3\_man\_in\_forest\_Gender\_Beard.avi}
&\multicolumn{1}{l}{(d) 4\_man\_with\_muscle\_Smiling\_Young.avi}
\\
\multicolumn{1}{l}{\quad Gender, - Beard}
&\multicolumn{1}{l}{\quad + Smile, - Age}
\\
\includegraphics[width=\linewidth]{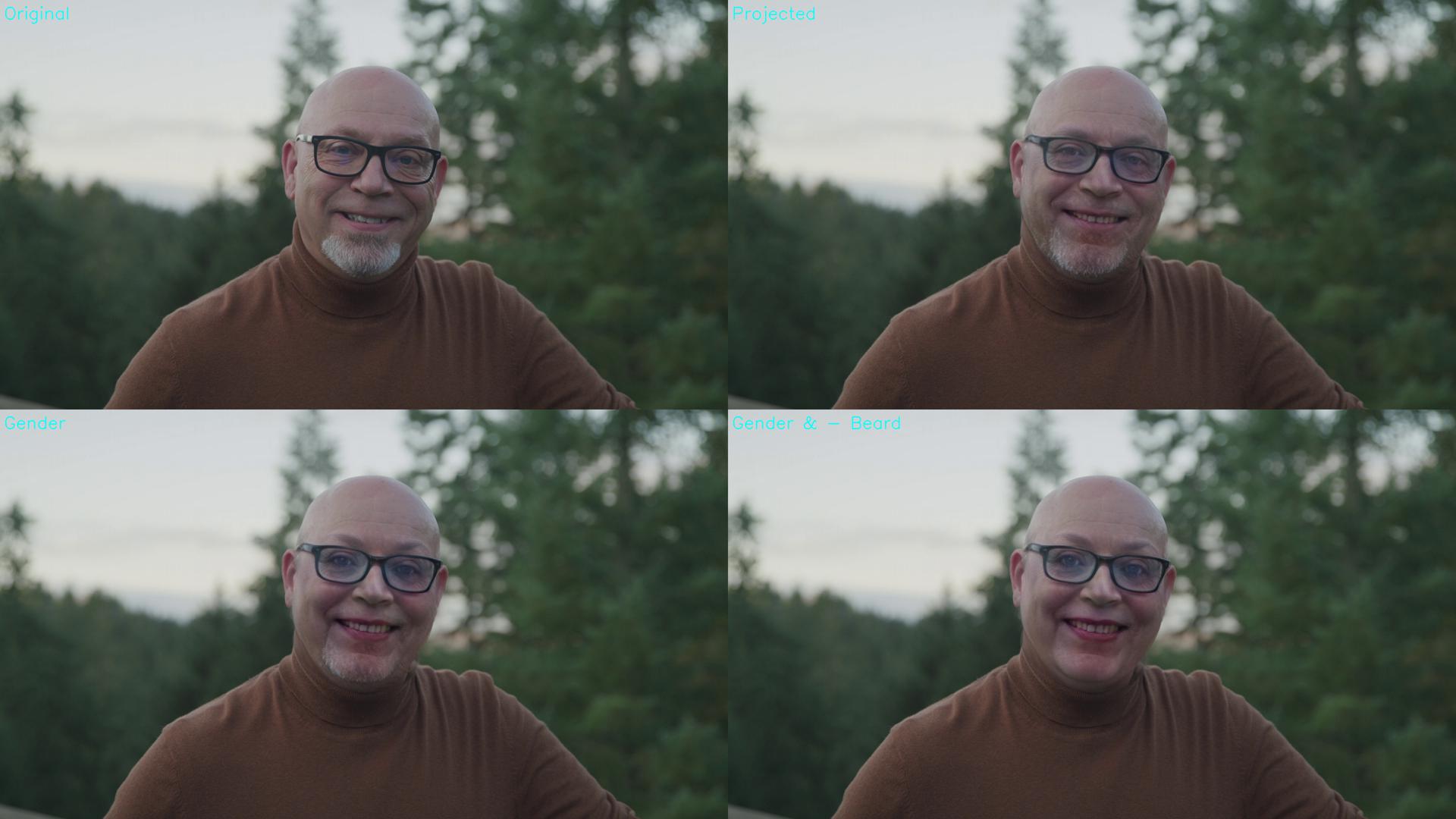}
&\includegraphics[width=\linewidth]{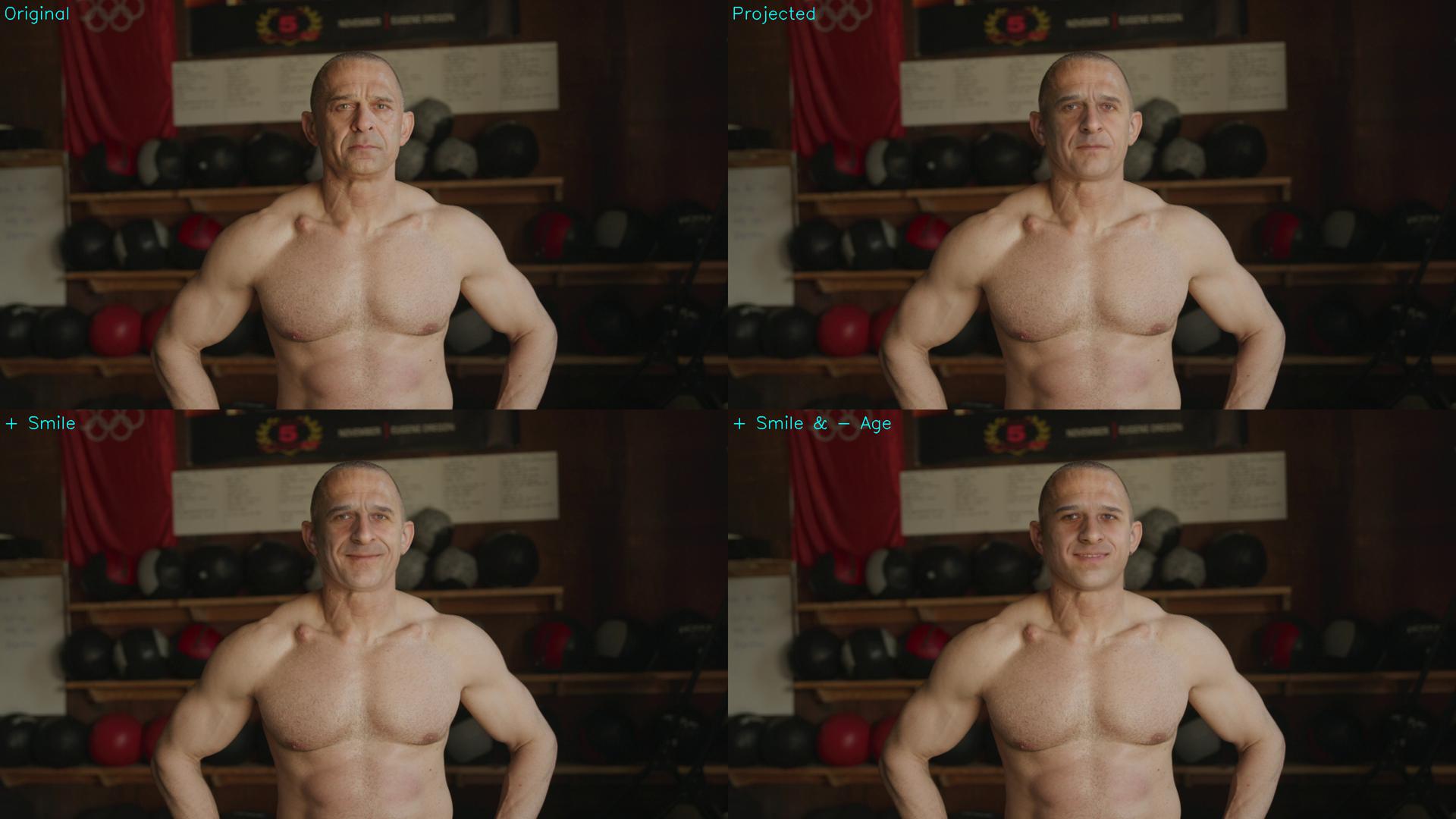}
\\
\hline
\\
\multicolumn{1}{l}{(e) 5\_man\_talking\_Bags\_Under\_Eyes\_Eyeglasses.avi}
&\multicolumn{1}{l}{(f) 6\_woman\_turning\_Smiling\_Makeup.avi}
\\
\multicolumn{1}{l}{\quad - Bags under eyes, + Eyeglasses}
&\multicolumn{1}{l}{\quad - Smile, + Makeup}
\\
\includegraphics[width=\linewidth]{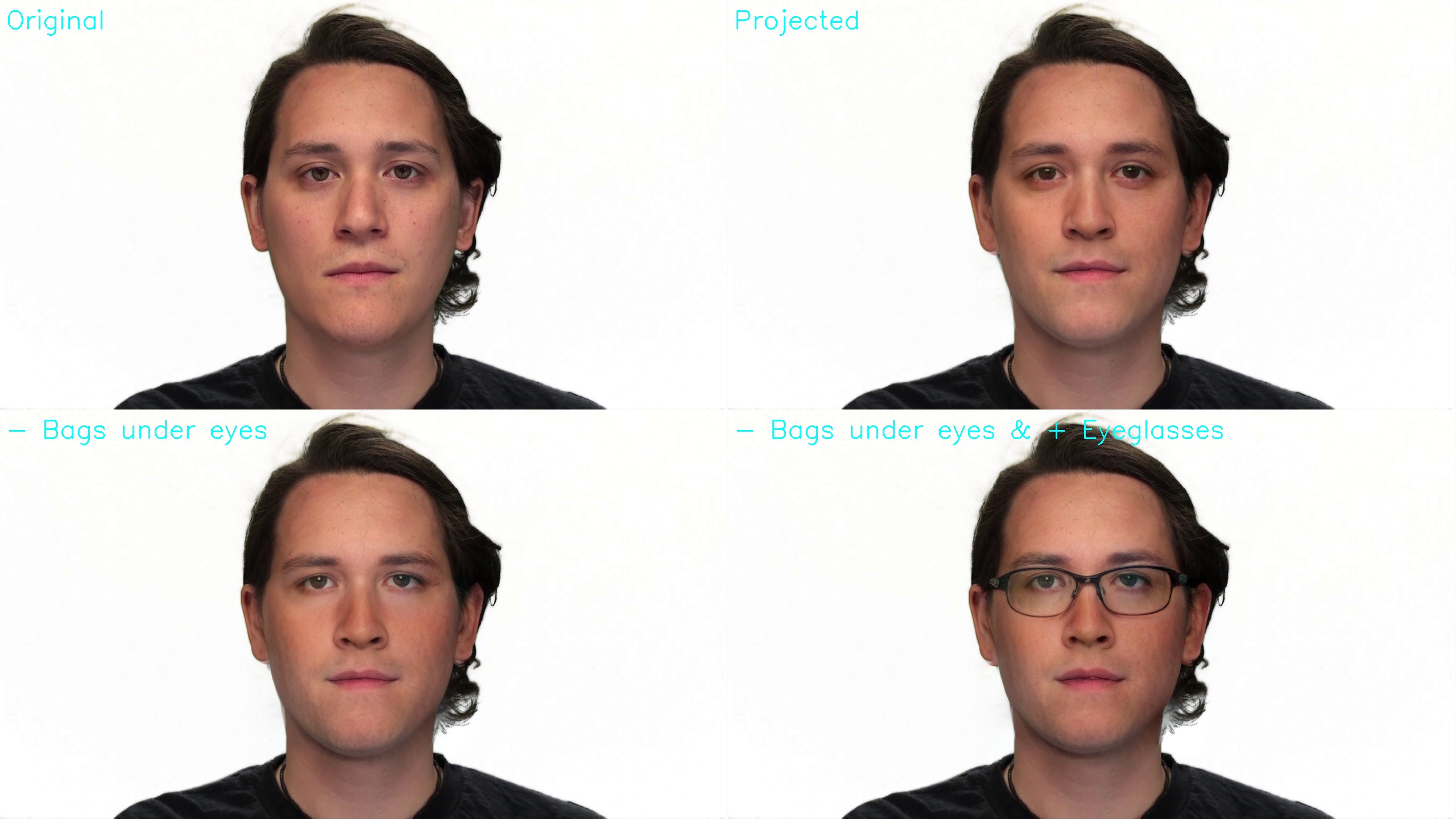}
&\includegraphics[width=\linewidth]{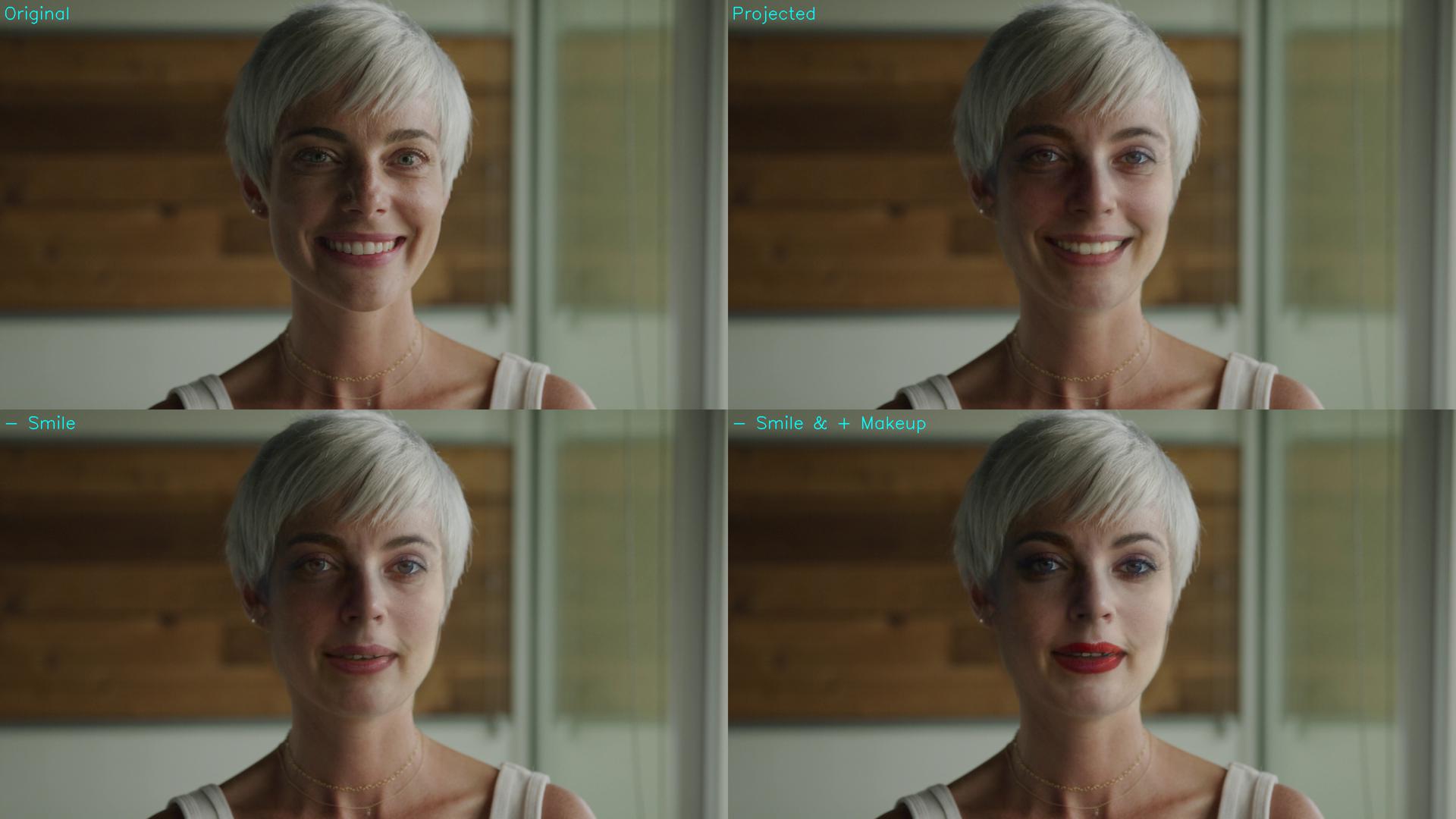}
\\
\end{tabular}
\caption{\textbf{Facial attribute editing on videos.} Each sub-figure corresponds to a frame extracted from the specified video, corresponding to the manipulation result of the indicated attributes. In each sub-figure, the upper row shows the original frame and the projected frame reconstructed with the encoded latent code in StyleGAN, the bottom row shows the manipulated frames for the first attribute and then for two attributes. Please open the video files to visualize the manipulation details.
}
\label{sup:video_fig}
\end{figure*}

\begin{figure*}[t]
\centering
\small
\setlength{\tabcolsep}{1pt}
\renewcommand{\arraystretch}{1}
\begin{tabular}{c c}
\\
\multicolumn{2}{l}{7\_woman\_with\_bricks\_progressive\_Smiling.avi, + Smile progressively}
\\
\multicolumn{2}{c}{\includegraphics[width=0.98\linewidth]{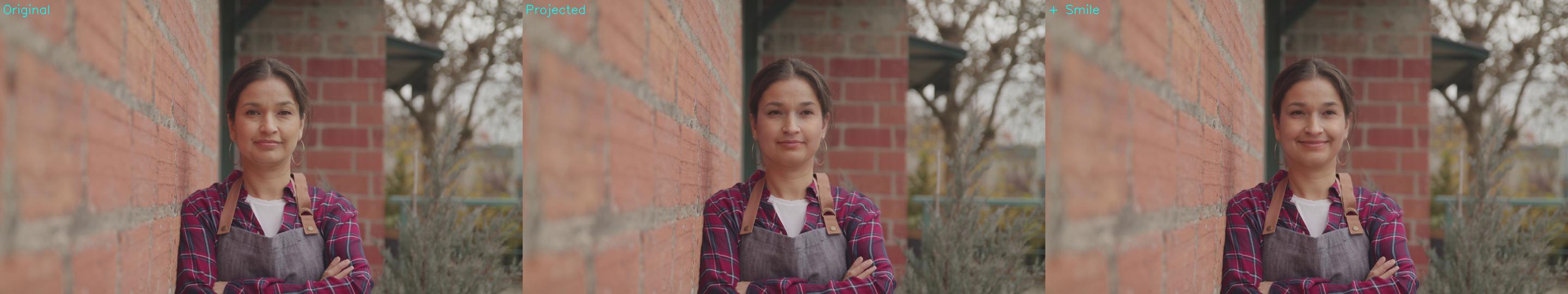}}
\end{tabular}
\caption{\textbf{Progressive attribute editing on videos.} 
By varying the scaling factor progressively along the sequence, the corresponding attribute is gradually varied. This figure show a frame extracted from the edited video, which corresponds to the progressive manipulation of the attribute `smile'. From left to right: the original frame, the projected frame, and the manipulated frame. Please open the video file to fully visualize the manipulation.
}
\label{sup:progress}
\end{figure*}

\begin{figure*}[t]
\centering
\small
\setlength{\tabcolsep}{0pt}
\renewcommand{\arraystretch}{0.5}
\begin{tabular}{P{0.16\linewidth}P{0.16\linewidth}P{0.16\linewidth}}
\multicolumn{3}{l}{failure\_case\_woman\_sitting\_Makeup.avi, + Makeup}
\\
\multicolumn{3}{c}{\includegraphics[width=0.48\linewidth]{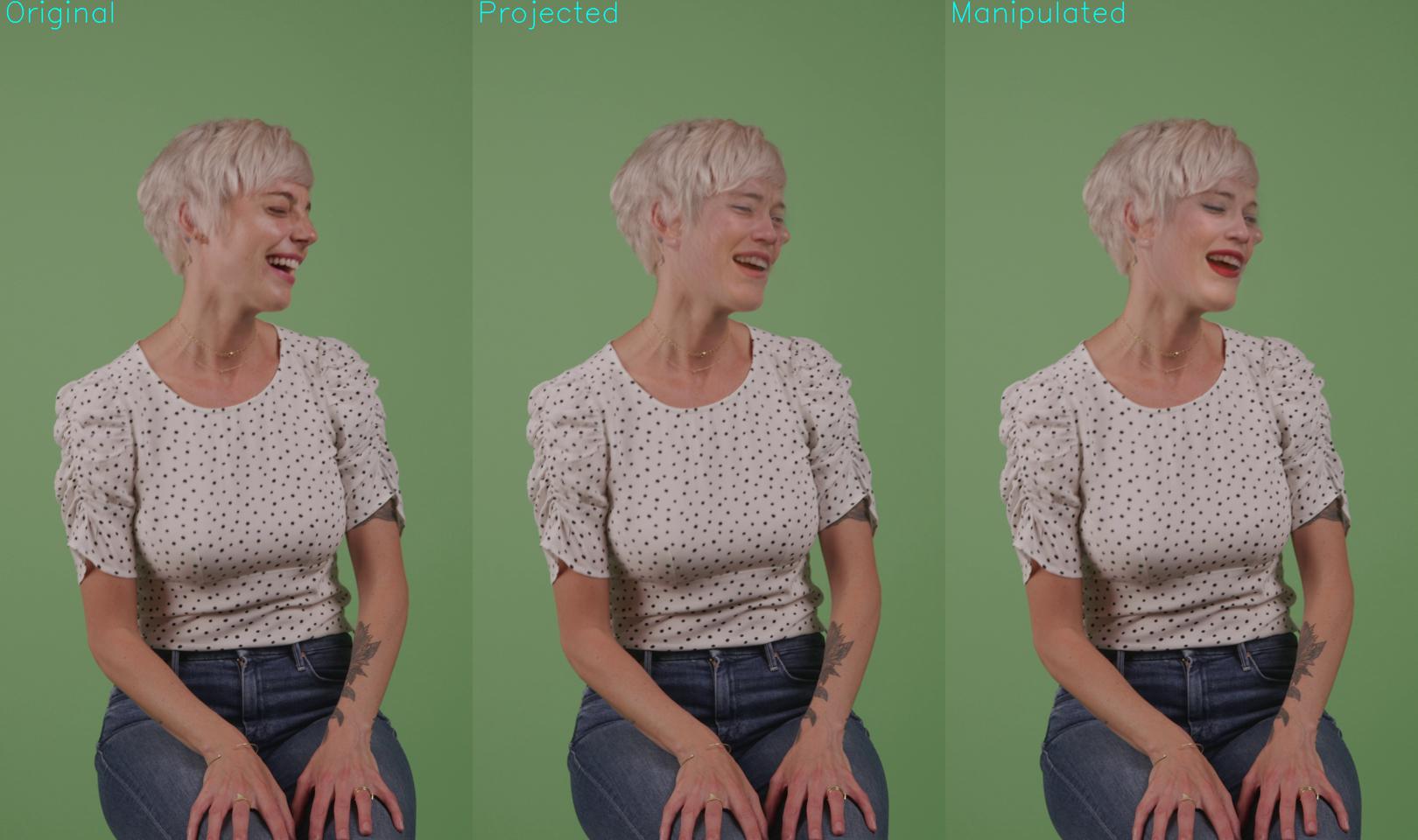}}
\\
\end{tabular}
\caption{Failure case of attribute manipulation on a video. This is a side pose frame extracted from the named video, which is the manipulation result of the attribute `makeup'. From left to right:  the original frame, the projected frame, and the manipulated frame. The face is not well reconstructed in the projected frame, and consequently the manipulated output contains defects. This is due to the limited generation capacity of the pre-trained encoder and the StyleGAN generator.
}
\label{sup:fail_vid}
\end{figure*}

As mentioned in Sec. 5.2 of the main paper, we present additional facial attribute editing results on videos in Figure \ref{sup:video_fig}. Each sub-figure corresponds to a frame extracted from the corresponding video, in which the indicated attributes are modified. For each video, we edit two attributes sequentially, and generate disentangled manipulation results. For example, in Figure \ref{sup:video_fig}(c) when changing the person to woman, our method does not influence the attribute `beard', despite the fact that it is correlated with gender. Besides, by varying the scaling factor progressively along the sequence, we achieve progressive attribute editing on videos. As shown by the video in Figure \ref{sup:progress}, we can simulate a progressive smiling process by smoothly varying the scaling factor. Overall, our method generates stable and consistent manipulation results on videos, provided that motion is not too strong. 
When there are quick changes of pose, we observe lighting or geometric artifacts. These artifacts are in fact due to the projection in the latent space, and therefore necessarily extend to the manipulated videos. As can be seen from the video in Figure \ref{sup:fail_vid}, the manipulation during the first half of the video is realistic and consistent. But when the face turns to a side pose, the projected face is not well reconstructed and therefore neither is the manipulated face. This may be due to the limited reconstruction capacity of the pre-trained encoder and StyleGAN model when the pose is not frontal.


\end{document}